\documentclass[10pt,journal,compsoc]{IEEEtran}

\ifCLASSOPTIONcompsoc
  \usepackage[nocompress]{cite}
\else
  \usepackage{cite}
\fi
\ifCLASSINFOpdf
\else
\fi
\usepackage{multirow}
\usepackage{booktabs}
\usepackage{tabularx}

\usepackage{amsfonts}  
\usepackage{amsmath}  

\usepackage{amsthm}  

\usepackage{graphicx} 
\usepackage{subfigure} 

\usepackage{hyperref}  

\begin{document}
\title{Dynamic Feature Learning and Matching for Class-Incremental Learning}  
\author{Sunyuan~Qiang, Yanyan~Liang, Jun~Wan,  and~Du~Zhang 
\IEEEcompsocitemizethanks{\IEEEcompsocthanksitem S. Qiang, Y. Liang, and D. Zhang are with Macau  University of Science and Technology (M.U.S.T), Macau 999078, China, 999078 (E-mail: 3220004460@student.must.edu.mo, yyliang@must.edu.mo,  duzhang@must.edu.mo).\protect \\ 
\IEEEcompsocthanksitem  
J. Wan is with the State Key Laboratory of Multimodal Artificial Intelligence Systems (MAIS), Institute of Automation, Chinese Academy of Sciences (CASIA), Beijing 100190, China, also with the School of Artificial Intelligence, University of Chinese Academy of Sciences (UCAS), Beijing 100049, China, and also with Macau University of Science and Technology (M.U.S.T), Macau  999078, China (e-mail: jun.wan@ia.ac.cn).  \protect \\}
\thanks{
Manuscript received X X, 2023; revised X X, 2023.
Corresponding author: Yanyan Liang.  
}}   

%
%

\markboth{Journal of \LaTeX\ Class Files,~Vol.~14, No.~8, August~2015}%
{Shell \MakeLowercase{\textit{et al.}}: Bare Advanced Demo of IEEEtran.cls for IEEE Computer Society Journals}
%



\IEEEtitleabstractindextext{%
\begin{abstract}

Class-incremental learning (CIL) has emerged as a means to learn new classes incrementally without catastrophic forgetting of previous classes. Recently, CIL has undergone a paradigm shift towards dynamic architectures due to their superior performance. However, these models are still limited by the following aspects: (\romannumeral1) \textit{Data augmentation (DA).} DA strategies, which are tightly coupled with CIL, remains under-explored in dynamic architecture scenarios. (\romannumeral2) \textit{Feature representation.} The discriminativeness of dynamic feature are sub-optimal and possess potential for refinement. (\romannumeral3)  
\textit{Classifier.} The misalignment between dynamic feature and classifier constrains the capabilities of the model. To tackle the aforementioned drawbacks, we propose the \textit{Dynamic Feature Learning and Matching} (DFLM) model in this paper from above three perspectives.  Specifically, we firstly introduce class weight information and non-stationary functions to extend the mix DA method for dynamically adjusting the focus on memory during training. Then, von Mises-Fisher (vMF) classifier is employed to effectively model the dynamic feature distribution and implicitly learn their discriminative properties. Finally, the matching loss is proposed to facilitate the alignment between the learned dynamic features and the classifier by minimizing the distribution distance. Extensive experiments on CIL benchmarks validate that our proposed model achieves significant performance improvements over existing methods.

 
\end{abstract}

}

\maketitle

\IEEEdisplaynontitleabstractindextext

%
\IEEEpeerreviewmaketitle

\ifCLASSOPTIONcompsoc
\IEEEraisesectionheading{\section{Introduction}\label{sec:introduction}}
\else
\section{Introduction}
\label{sec:introduction}
\fi

%
%
%
%
\IEEEPARstart{T}{he} complex and ever-changing nature of real-world environments necessitates that deep neural networks (DNNs) possess the ability to constantly acquire novel knowledge and adapt to new circumstances, especially in areas like computer vision~\cite{DBLP:journals/pami/JosephRKKB22,DBLP:conf/cvpr/RebuffiKSL17}, natural language processing~\cite{DBLP:journals/corr/abs-2211-12701,DBLP:journals/corr/abs-1802-06024,DBLP:conf/nips/Ouyang0JAWMZASR22}, and autonomous vehicles~\cite{DBLP:journals/tits/HanZCQQLZ23,DBLP:conf/cvpr/MirzaMPB22}, where data can exhibit rapid and unpredictable variations. Despite the remarkable achievements~\cite{DBLP:journals/pami/GirshickDDM16,DBLP:conf/cvpr/HeZRS16,DBLP:conf/nips/KrizhevskySH12,DBLP:journals/corr/SimonyanZ14a} of neural networks in static scenarios, catastrophic forgetting remains a major challenge~\cite{french1999catastrophic,DBLP:journals/corr/GoodfellowMDCB13,MCCLOSKEY1989109} for their application in dynamic real-world environments, where models exhibit a bias towards adapting exclusively to the new task data distribution, while disregarding the knowledge acquired from previous tasks.  
To mitigate this issue, the class-incremental learning (CIL) paradigm is introduced to enable models to continually incorporate discriminative representations of newly arrived categories while maintaining knowledge learned from previously tasks, which has garnered significant attention in recent academic community~\cite{DBLP:journals/nn/BelouadahPK21,DBLP:journals/pami/LangeAMPJLST22,DBLP:conf/eccv/DouillardCORV20,DBLP:conf/cvpr/class2022kang,DBLP:conf/nips/LiuSS21,DBLP:conf/cvpr/RebuffiKSL17,DBLP:conf/cvpr/YanX021,DBLP:conf/cvpr/ZhaoXGZX20,DBLP:conf/nips/ZhuCZL21,DBLP:conf/eccv/WangZYZ22}.

The Class-incremental learning  approaches can be primarily divided into three directions: regularization methods, memory replay methods, and dynamic architecture-based methods. Additionally, the combination of these directions has resulted in the emergence of more powerful algorithms. A common approach for achieving continual knowledge updating in the early works is based on \textit{regularization} strategies~\cite{DBLP:conf/eccv/AljundiBERT18,DBLP:journals/corr/KirkpatrickPRVD16,DBLP:conf/eccv/LiH16}.  
These methods aimed to alleviate the issue of catastrophic forgetting by introducing additional constraints of the model parameters in the training process. As for \textit{memory replay} methods~\cite{DBLP:journals/corr/abs-1902-10486,DBLP:conf/cvpr/RebuffiKSL17,DBLP:conf/nips/ShinLKK17}, they mainly alleviate forgetting of past tasks by jointly training on the stored (or synthesized) old examples and current new task samples. The \textit{dynamic architecture-based} methods~\cite{DBLP:conf/nips/RajasegaranH0K019,DBLP:journals/corr/RusuRDSKKPH16,DBLP:conf/cvpr/YanX021} usually allocates extra model parameter space for learning new task knowledge without compromising the performance of old tasks. Moreover, current exemplar replay techniques are also widely integrated with knowledge distillation (KD)-based regularization methods \cite{DBLP:conf/eccv/DouillardCORV20,DBLP:journals/corr/HintonVD15,DBLP:conf/cvpr/class2022kang,DBLP:conf/eccv/LiH16} and dynamic architecture models   \cite{DBLP:conf/cvpr/YanX021,DBLP:conf/eccv/WangZYZ22} to further improve performance in CIL. Among them, memory replay methods~\cite{DBLP:conf/cvpr/RebuffiKSL17} with dynamic architecture are widely utilized in the CIL community owing to their promising performance.

However, despite the advancements made in these models, their effectiveness is still suffers from certain limitations in the following aspects: (\romannumeral1) In dynamic architectures, data augmentation strategies are still relatively under-explored. We study widely used powerful mix-based DA methods, namely,  Mixup~\cite{DBLP:conf/iclr/ZhangCDL18} and CutMix~\cite{DBLP:conf/iccv/YunHCOYC19}, but our empirical results presented in Table~\ref{table_mix_based_data_aug} suggest that directly transferring the vanilla mixed strategies to the dynamic CIL does not yield substantial improvements. The global static interpolation coefficient $\lambda$ distribution over epoch is not suitable for the case of insufficient memory samples in the CIL scenarios.  
(\romannumeral2) The concatenated dynamic feature representation falls into the sub-optimal situations that restrict the performance of the model. Fig.~\ref{plot_tsne_der_and_ours} visually illustrates the partial overlap of features. The nearest-mean-of-exemplars (NME)~\cite{DBLP:conf/cvpr/RebuffiKSL17}, as a predictor based on class feature
prototypes, can reflect the quality of features. Quantitatively, there is still a room for improvement shown in Table~\ref{table_main_ablation_comparision} compared to CNN. (\romannumeral3) The phenomenon of mismatch between classifier and features, as observed through the CNN and NME accuracies in Table~\ref{table_main_ablation_comparision}, highlights the inability to accurately model the distribution of obtained dynamic features, leading to a degradation in model prediction. And the comprehensive analysis is provided in ablation study in section~\ref{section_ablation_study}.

To this end, we propose the Dynamic Feature Learning and
Matching (DFLM) model in this paper. In more detail, we firstly introduce the Memory-Centric Mix (MC-Mix) DA strategy, which serves as an extended variant of CutMix, not only incorporates the regularization capability of the mixed DA, but also further integrates class weight information and non-stationary functions to dynamically adjust the linear interpolation coefficient $\lambda$ in the label space, which enables the model to dynamically focus on memory samples during the training process. The experimental results in  Table~\ref{table_mix_based_data_aug} show that the best performance gain can be obtained by our proposed MC-Mix in this paper. Secondly, the vMF classifier is utilized for modeling the distribution of dynamic features without the need for additional parameter estimation, while also implicitly learning discriminative properties based on the classification loss. Finally, we build the distribution of batch-wise feature  prototypes using the above vMF classifier modeling and align them with the classifier weights by optimizing the Kullback-Leibler (KL) divergence from the distribution perspective. Extensive experiments on CIFAR-100, ImageNet-100, and
ImageNet-1000 datasets with a total of 13 different evaluation settings validate that our model exhibits superior performance on CIL benchmarks. In particular, we achieve about $76 \%$ and $79 \%$ average accuracy on CIFAR-100 and ImageNet100 B0 10 steps benchmarks, respectively. 
 
The main contributions of our work are summarized as follows: (1) We propose a novel mixed data augmentation (DA) strategy in dynamic CIL scenario. (2) We introduce to model and learn the dynamic features with vMF classifier. (3) We propose a dynamic matching loss to align the vMF classifier with dynamic  features. (4) We achieve superior results on three
commonly used datasets with 13 different evaluation settings in CIL.

\section{Related Work}
\label{section_related_work}

In this section, we provide a  comprehensive overview of the existing literature related to our proposed method, which we organize into three modules: class-incremental learning (CIL), data augmentation, and cosine classifier. 
 
\subsection{Class-Incremental Learning}


There are three main directions~\cite{DBLP:journals/nn/BelouadahPK21,DBLP:journals/corr/abs-2302-03648,DBLP:journals/corr/abs-2302-00487,DBLP:journals/pami/LangeAMPJLST22,DBLP:journals/pami/MasanaLTMBW23} of class-incremental learning methods: regularization methods, memory replay methods, and dynamic architecture-based methods. Furthermore, combining these directions has led to the development of more powerful algorithms. In this subsection, we comprehensively discuss the methods in each category.

\textit{Regularization} methods~\cite{DBLP:conf/eccv/AljundiBERT18,DBLP:journals/corr/KirkpatrickPRVD16,DBLP:conf/eccv/LiH16,DBLP:conf/icml/ZenkePG17} aim to prevent models from overfitting to new classes and forgetting old ones by applying regularization techniques to the model parameters. 
Notably, LwF~\cite{DBLP:conf/eccv/LiH16} firstly used knowledge distillation  (KD)~\cite{DBLP:journals/corr/HintonVD15} in continual learning, which subsequently greatly inspired the CIL community and is often combined with replay exemplars (exemplar-based KD methods, EKD)~\cite{DBLP:conf/cvpr/RebuffiKSL17,DBLP:conf/eccv/DouillardCORV20}. 
 
\textit{Memory replay} methods~\cite{DBLP:journals/corr/abs-1902-10486,DBLP:conf/cvpr/RebuffiKSL17,DBLP:conf/nips/Lopez-PazR17,DBLP:conf/nips/ShinLKK17,DBLP:conf/nips/GoodfellowPMXWOCB14,DBLP:journals/corr/KingmaW13,van2020brain} involve either directly storing a small amount of old data or using a neural network model to synthesize old data for training, often referred to as memory exemplar replay and generative replay, respectively. Deep generative replay models typically require a large parameter space and may produce images of varying quality for training. 
For directly saving exemplars~\cite{DBLP:conf/nips/Lopez-PazR17,DBLP:conf/cvpr/RebuffiKSL17},  
iCaRL~\cite{DBLP:conf/cvpr/RebuffiKSL17} provided two valuable insights to the CIL community: using KD-based regularization methods~\cite{DBLP:conf/eccv/LiH16,DBLP:journals/corr/HintonVD15} and maintaining exemplar sets can effectively alleviate the forgetting problem. Later, various exemplar-based KD methods (EKD) 
emerged~\cite{DBLP:conf/eccv/CastroMGSA18,DBLP:conf/eccv/DouillardCORV20,DBLP:conf/cvpr/class2022kang}.   
Moreover, 
due to the tiny number of memory samples, some CIL models mitigate catastrophic forgetting from an imbalance  view~\cite{DBLP:conf/eccv/CastroMGSA18,DBLP:conf/cvpr/HouPLWL19,DBLP:conf/cvpr/WuCWYLGF19,DBLP:conf/cvpr/ZhaoXGZX20,qiang2023mixture}.      
Some researchers further shifted their focus back to the original strict non-exemplar~\cite{DBLP:conf/cvpr/0004TLHWCJ020,DBLP:conf/cvpr/ZhuZWYL21,DBLP:conf/nips/ZhuCZL21,DBLP:conf/cvpr/0004Z0LZ22} and data-free settings~\cite{DBLP:conf/cvpr/YinMALMHJK20,DBLP:conf/iccv/SmithHBSJK21,DBLP:conf/eccv/GaoZGZ22}.
On the other hand, some works~\cite{DBLP:conf/nips/LiuSS21,DBLP:journals/corr/abs-2205-13218,DBLP:conf/iclr/WangZYYLHZLZZ22} centered their attention on the selection of replay exemplars.
Different from our paper, we consider the scenario of incorporating exemplar sets with dynamic architecture-based models.


As for \textit{Dynamic architecture-based} methods~\cite{DBLP:conf/nips/RajasegaranH0K019,DBLP:journals/corr/RusuRDSKKPH16,DBLP:conf/cvpr/YanX021,DBLP:conf/eccv/WangZYZ22}, they usually open up additional parameter space to learn new tasks and exhibit state-of-the-art performance. 
Despite the drawback of in terms of the parameter space, this paper focuses on exploring the factors that limit the performance potential of dynamic architecture models.  Recently, some research~\cite{DBLP:conf/cvpr/WuSLRVBS22,DBLP:journals/csur/LiuYFJHN23,DBLP:conf/cvpr/0002ZL0SRSPDP22,DBLP:conf/eccv/0002ZESZLRSPDP22,DBLP:conf/iclr/DosovitskiyB0WZ21} efforts have migrated the setup of continual learning to pre-trained models and design additional tunable prompts to facilitate the model's ability for continual learning.  
However, 
this paper focus on exploring the dynamic architecture-based models without any pre-trained models to further improve performance under CIL benchmarks.

\subsection{Data Augmentation in CIL}
\label{section_related_work_da_in_cil}

\begin{figure*}[t]
\centering 
\includegraphics[width=0.8\textwidth]{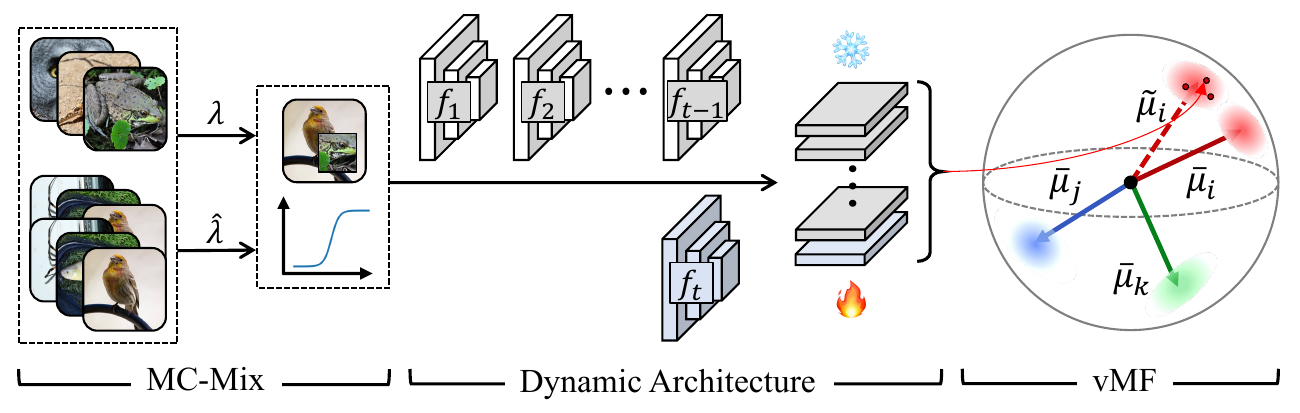}  
\caption{The overview of the proposed model. }  
\label{fig_overview}
\end{figure*}

In CIL, the issue of catastrophic forgetting can arise due to the absence of past task data. In this scenario, data augmentation (DA)~\cite{DBLP:journals/jbd/ShortenK19,DBLP:journals/pr/XuYFP23} becomes tightly intertwined with CIL. 
\textit{Model-free} DA  methods~\cite{DBLP:conf/aaai/Zhong0KL020,DBLP:journals/corr/abs-1708-04552,DBLP:conf/iclr/ZhangCDL18,DBLP:conf/iccv/YunHCOYC19,DBLP:conf/icml/HanFLHARPL22} involve directly manipulating images through various  transformations, which are usually plug-and-play in deep learning model training. 
Additionally, mix-based DA techniques~\cite{DBLP:conf/iclr/ZhangCDL18,DBLP:conf/icml/VermaLBNMLB19,DBLP:conf/iccv/YunHCOYC19} generate new virtual samples by linearly interpolating the pixels, features, and labels of two training samples, which further regularize model training and improve performance. Given the universal plug-and-play nature of the model-free DA methods, they were seamlessly used in CIL and brought significant performance improvements~\cite{DBLP:conf/eccv/CastroMGSA18,DBLP:conf/cvpr/BangKY0C21,DBLP:conf/cvpr/ZhuZWYL21,DBLP:conf/nips/ZhuCZL21,More2022ZhuClass,qiang2023mixture}. 
\textit{Model-based} DA methods~\cite{DBLP:journals/corr/abs-1711-04340,DBLP:conf/cvpr/JacksonABBO19} typically rely on powerful deep generative models~\cite{DBLP:conf/nips/GoodfellowPMXWOCB14,DBLP:journals/corr/RadfordMC15} to synthesize virtual data for model training, 
which can be directly categorized as generative replay-based 
methods~\cite{DBLP:conf/nips/ShinLKK17,van2020brain} in CIL.  
\textit{Optimizing policy-based} DA methods~\cite{DBLP:conf/cvpr/CubukZMVL19,DBLP:conf/cvpr/CubukZSL20} automatically search for the optimal data augmentation policies for a given dataset and task. 
In CIL, these DA  methods are also used to resist forgetting~\cite{DBLP:conf/cvpr/BangKY0C21,DBLP:conf/eccv/WangZYZ22}. In this paper, we additionally add this strategy and observe a performance improvement.  
For \textit{implicit based} DA methods~\cite{DBLP:conf/nips/WangPSZHW19,DBLP:journals/pami/WangHSPXW22,gao2022hyperbolic}, they aim to implicitly transform the data augmentation into an optimized form of the loss function.  
In CIL, semantic augmentation~\cite{DBLP:conf/nips/ZhuCZL21}  directly apply ISDA strategy on each old class using preserved previous class statistics. 
However, these methods typically require estimation of parameters for the assumed distribution, 
and the quality of parameter estimation greatly affects the effectiveness of data augmentation~\cite{DBLP:conf/cvpr/LiGLWQC21}.

\subsection{Cosine Classifier in CIL}

The cosine classifier~\cite{DBLP:conf/icann/LuoZXWRY18} is commonly used in deep learning for tasks like face recognition~\cite{DBLP:conf/cvpr/LiuWYLRS17,DBLP:conf/cvpr/WangWZJGZL018,DBLP:conf/cvpr/DengGXZ19}, long-tail recognition~\cite{DBLP:conf/bmvc/IscenAGS21,DBLP:conf/cvpr/LiCL22,DBLP:conf/eccv/WangFHFLH22,DBLP:conf/cvpr/AlshammariWRK22}, and class-incremental learning~\cite{DBLP:conf/cvpr/HouPLWL19,DBLP:conf/eccv/DouillardCORV20}. Furthermore, the cosine classifier exhibits a natural suitability for modeling with the von Mises-Fisher (vMF)  distribution~\cite{DBLP:conf/eccv/WangFHFLH22,DBLP:journals/corr/HasnatBMGC17,DBLP:conf/iccv/ScottGM21,DBLP:conf/cvpr/LiXXSLH21,DBLP:journals/pr/ZheCY19,DBLP:conf/eccv/KirchhofRAK22,DBLP:conf/icml/0001I20,DBLP:conf/icml/ZimmermannSSBB21,DBLP:conf/cvpr/000121}, which confers upon it a property  characterized by a parameterized distribution description.   
In CIL,  UCIR~\cite{DBLP:conf/cvpr/HouPLWL19} uses a cosine classifier to alleviate model bias towards new class predictions in incremental steps. PODNet~\cite{DBLP:conf/eccv/DouillardCORV20} proposes a local similarity classifier (LSC) to consider multiple proxies per class of cosine classifiers. However, the analysis of the cosine classifier using vMF modeling has been overlooked in these CIL methods, while this paper further explores the application of vMF modeling methods in CIL. 
 
\section{Method}

In this section, we first introduce the problem definition of CIL and dynamic architecture-based strategy, then describe our proposed dynamic feature learning and matching method, which mainly consists of three components, memory-centric mix, vMF classifier, and dynamic feature matching loss. We provide an overview of the proposed method in the final subsection and the framework is depicted in Fig.~\ref{fig_overview}.       

\subsection{Problem Definition}
\label{subsection_problem_definition}

In class-incremental learning paradigm~\cite{DBLP:conf/cvpr/RebuffiKSL17}, the datasets for each task are sequentially received over time. Specifically, the training dataset $\mathcal{D}_t$ at each step $t$ is composed of $N_t$ sample pairs denoted by $\{(\mathbf{x}_{t,i}, y_{t,i})\} _{i=1}^{N_t}$, where $\mathbf{x}_{t,i}$ and $y_{t,i}$ are the $i^\text{th}$ sample pair at the $t^\text{th}$ step from data and target space $\mathcal{X}_t$ and $\mathcal{Y}_t$, respectively. The CIL setting assumes that the target spaces at different steps are non-overlapping, i.e., $\mathcal{Y}_i \cap \mathcal{Y}_j = \emptyset$ for $i \not= j$. Particularly, only a small subset of the previous training dataset is retained to construct the exemplar set $\mathcal{M}_t  \subset \mathcal{D}_{1:t-1} = \cup_{i=1}^{t-1}\mathcal{D}_i$ with a fixed size $K$. At step $t$, the model is trained on available dataset $\mathcal{D}_t \cup \mathcal{M}_t$ and required to make predictions across all seen classes $\mathcal{Y}_{1:t}=\cup_{i=1}^t \mathcal{Y}_i$. $|\mathcal{Y}_{1:t-1}|$, $|\mathcal{Y}_{t}|$, and $|\mathcal{Y}_{1:t}|$ denote the total number of old classes, new classes, and all classes, respectively.

\subsection{Dynamic Architecture}
\label{section_dynamic_architecture}
  
Dynamic architecture-based strategy~\cite{DBLP:conf/cvpr/YanX021,DBLP:conf/eccv/WangZYZ22} is   adopted in this paper. An additional feature extractor is allocated and the previous feature extractors are frozen to learn the knowledge of the new coming categories while maintaining discriminative features of past categories. Formally, as stated in section~\ref{subsection_problem_definition}, at each task step $t$, the expanded feature extractor $f_t$ is concatenated with the previously fixed accumulated extractors $\mathcal{F}_{t-1}$ to obtain a new joint extractor  $\mathcal{F}_t$.  
\begin{equation}
    \mathbf{z} = \mathcal{F}_t(\mathbf{x}) = [ f_t(\mathbf{x}) ; \mathcal{F}_{t-1}(\mathbf{x}) ], 
\label{equation_dynamic_architecture}
\end{equation} 
where $\mathbf{z} \in \mathbb{R}^{d_t}$ denotes the feature extracted by $\mathcal{F}_t$ with dimension $d_t$ at step $t$, and $[ \cdot \,; \cdot ]$ denotes the concatenate operation. Then, a fully connected layer is followed for classification $\mathbf{W}_t^\top \mathbf{z}$, where $\mathbf{W}_t \in \mathbb{R}^{d_t \times |\mathcal{Y}_{1:t}|}$ is the weight matrix. The cross-entropy loss function is commonly employed for model training.  
\begin{equation}      
     \mathcal{L}_\mathrm{CE} =  - \sum^{C} y  \cdot  \log( p ( \mathbf{z} )  )   , 
\label{equation_cross_entropy_default_loss_function}
\end{equation}   
where $p ( \mathbf{z}) = \mathrm{softmax} \!  \left( \mathbf{W}_t^\top \mathbf{z}  \right)$ is the output probability of all seen classes $|\mathcal{Y}_{1:t}|$. Moreover, an auxiliary fully connected layer~\cite{DBLP:conf/cvpr/YanX021} is often introduced into the model to distinguish between old and new classes, where the previous old classes are treated as a unified one class, i.e. $\mathbf{W}^{\mathrm{Aux}}_t \in \mathbb{R}^{d_t \times \left(1+ |\mathcal{Y}_{t}| \right) }$ is the auxiliary weight matrix.  
\begin{equation}      
     \mathcal{L}_\mathrm{Aux} =  - \sum^{C} \hat{y} \cdot  \log( \hat{p} ( \mathbf{z} )  )   , 
\label{equation_aux_loss_function}
\end{equation}       
where $\hat{y}$ denotes a new label assignment based on $y$ where all classes of the old tasks are grouped into one class, while the remaining classes of the new task are kept same. $\hat{p} (\mathbf{z}) = \mathrm{softmax} \!  \left( {\mathbf{W}^{\mathrm{Aux}}_t}^\top  \mathbf{z} \right)$ is the output probability of new assignment classes $1 + |\mathcal{Y}_{t}|$. In the following sections, we omit the task index symbol $t$ for  simplicity.          
 
\subsection{Memory-Centric Mix} 
\label{section_memory_centric_mix}

Given the close relationship between data augmentation (DA) and CIL as discussed in section~\ref{section_related_work_da_in_cil}, our aim is to leverage mixed-based DA to learn better dynamic features and alleviate forgetting. In this subsection, we first discuss the existing two classic mix-based DA techniques~\cite{DBLP:conf/iclr/ZhangCDL18,DBLP:conf/iccv/YunHCOYC19}, and then introduce the proposed Memory-Centric Mix.     
  

\textbf{Mixup and Cutmix.} 
Mixup~\cite{DBLP:conf/iclr/ZhangCDL18} is the first mix-based DA technique that extends the empirical risk minimization (ERM)~\cite{DBLP:conf/nips/Vapnik91} to vicinal risk minimization (VRM)~\cite{DBLP:conf/nips/ChapelleWBV00} by a linear interpolation between two randomly selected samples. 

\begin{equation}
    \begin{aligned} 
        \tilde{\mathbf{x}}_{i,j} & = \lambda \cdot \mathbf{x}_i + (1 - \lambda) \cdot {\mathbf{x}_j} ,\\
        \tilde{y}_{i,j} & = \lambda \cdot y_i + (1 - \lambda) \cdot {y_j} ,
    \end{aligned}
\label{equation_mixup}
\end{equation}
where $(\mathbf{x}_i,y_i)$ and $(\mathbf{x}_j,y_j)$ are two sample pairs from training data distribution. $\lambda$ is usually drawn from Beta distribution $\lambda \sim \mathrm{Beta}(\alpha,\alpha)$. 
Mixup is an effective technique for regularizing neural networks and improving performance in various applications~\cite{DBLP:conf/iclr/ZhangDKG021,DBLP:journals/corr/abs-2006-06049},
and has also led to the emergence of various mix-based DA methods~\cite{DBLP:conf/icml/VermaLBNMLB19,DBLP:conf/iccv/YunHCOYC19}.
Cutmix~\cite{DBLP:conf/iccv/YunHCOYC19} is one such mix-based DA method that builds upon Mixup's success by combining it with Cutout~\cite{DBLP:journals/corr/abs-1708-04552}, where a portion of an image is randomly removed and replaced with a patch from another image.
\begin{equation}
    \begin{aligned} 
    \tilde{\mathbf{x}}_{i,j} & = \mathbf{M} \odot \mathbf{x}_i + (\mathbf{1} - \mathbf{M} ) \odot {\mathbf{x}_j} ,\\
        \tilde{y}_{i,j} & = \lambda \cdot y_i + (1 - \lambda) \cdot {y_j} ,
    \end{aligned}
\label{equation_cutmix}
\end{equation}
where $\mathbf{M} \in \{0, 1\}^{W\times H}$ denotes a binary mask that specifies the region within the image that will be replaced, and $\odot$ denotes element-wise multiplication. Similarly, $(\mathbf{x}_i,y_i)$ and $(\mathbf{x}_j,y_j)$ are two sample pairs, $\lambda$ is sampled from a beta distribution $\lambda \sim \mathrm{Beta}(\alpha,\alpha)$
and is also utilized to represent the proportion of the replacement region's area with respect to the entire image.

\begin{figure*}[t]
\centering    
\subfigure[Sigmoid function.]{
    \begin{minipage}[t]{0.246\textwidth}
    \centering     \includegraphics[width=1\textwidth]{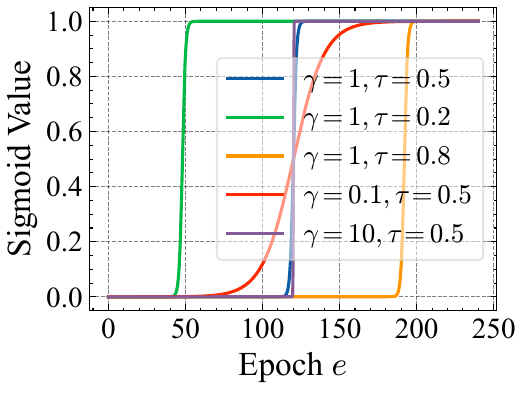} 
    \end{minipage} 
\label{figure_sigmoid_functions_gamma_tau}
}\hspace{-3mm}  
\subfigure[Mean function.]{
    \begin{minipage}[t]{0.246\textwidth}
    \centering
    \includegraphics[width=1\columnwidth]{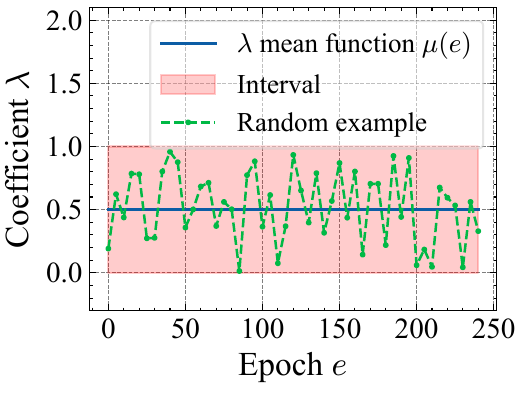} 
    \end{minipage}  
\label{figure_sigmoid_function_default_mean}  
}\hspace{-3mm}    
\subfigure[Mean function ($w_y=0.5$).]{
    \begin{minipage}[t]{0.246\textwidth}
    \centering  
    \includegraphics[width=1\columnwidth]{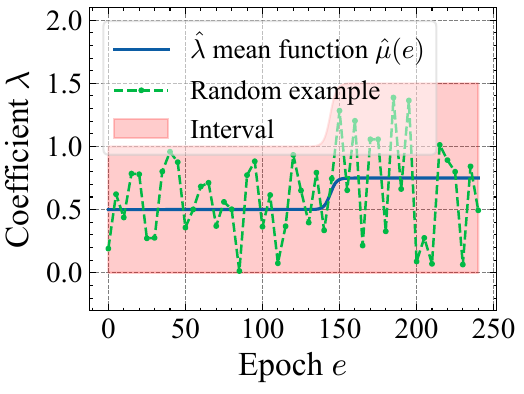} 
    \end{minipage}
\label{figure_sigmoid_function_old_weight}
}\hspace{-3mm} 
\subfigure[Mean function ($w_y=-0.5$).]{
    \begin{minipage}[t]{0.246\textwidth}  
    \centering
    \includegraphics[width=1\columnwidth]{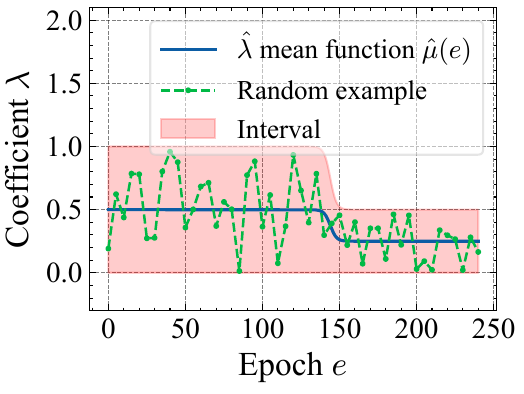} 
    \end{minipage}
\label{figure_sigmoid_function_new_weight}  
}  
\caption{(a) The visualization results of sigmoid function $\sigma_{\gamma,\tau}(e)$ with different $\gamma$ and $\tau$ in Eq.~\ref{equation_lambda_hat_rewrite}. (b)(c)(d) The visualization results of mean functions $\mu(e)$ and $\hat{\mu}(e)$ with intervals and random examples. }
\label{plot_sigmoid_function_mean_function}
\end{figure*}

\textbf{Memory-Centric Mix.} The Mixup and Cutmix strategies have shown strong performance in various applications, but their performance improvement in the CIL paradigm remains  limited. We conjecture that the potential of mix-based DA methods is severely limited in the incremental learning step due to the significant reduction in the memory replayed exemplars of past learned categories. Additionally, the abundance of newly coming task data may further suppress the contribution of old task data to the model's training. 


We first model the interpolation coefficient $\lambda$ as a time series process with respect to the training epoch $e$, where the total number of epochs is denoted as $E$. As shown in Eq.~\ref{equation_mixup} and Eq.~\ref{equation_cutmix}, the vanilla mix-based methods treat coefficient $\lambda(e) \sim \mathrm{Beta}(\alpha, \alpha)$ as a stationary sequence in both the sample and label spaces with the constant mean function as shown in Fig.~\ref{figure_sigmoid_function_default_mean}, 
   
\begin{equation}
    \mu(e) = \mathbb{E}_{\lambda(e)}  =  \frac{\alpha}{\alpha+\alpha} = \frac{1}{2}.
\label{equation_mean_function_constant}
\end{equation} 
Based on the Eq.~\ref{equation_mean_function_constant}, we argue that such a stationary sequence is not suitable for CIL:  
(\romannumeral1)  
Treating two samples to be mixed equally (with an expectation of $\frac{1}{2}$ each) ignores the issue of imbalance.
(\romannumeral2) The utilization of global static interpolation coefficients during training phase may potentially impair the performance of the model.
To this end,  we introduce a non-stationary time series process by extending the sigmoid function and incorporating the information of the class sample number to further improve the interpolation coefficients. According to the Cutmix~\cite{DBLP:conf/iccv/YunHCOYC19} and Eq.~\ref{equation_cutmix}, the mixed samples of our proposed memory-centric mix are also calculated by  
\begin{equation}
    \begin{aligned} 
    \tilde{\mathbf{x}}_{i,j} & = \mathbf{M} \odot \mathbf{x}_i + (\mathbf{1} - \mathbf{M} ) \odot {\mathbf{x}_j} , 
    \end{aligned}
\label{equation_cutmix_samples}
\end{equation} 
where $\mathbf{M} \in \{0, 1\}^{W\times H}$ denotes a binary mask corresponding to the coefficient $\lambda(e) \sim \mathrm{Beta}(\alpha,\alpha)$. $(\mathbf{x}_i,y_i)$ and $(\mathbf{x}_j,y_j)$ are two sample pairs. We maintain the original mix operation in the sample space, but for the label space, we extend $\lambda$ time series process with the sigmoid function $\sigma(x) = \frac{1}{1 + \exp(-x)}$,    
\begin{equation}    
    \begin{aligned}
    \hat{\lambda}(e, y, \lambda) & = \frac{w_y \cdot \lambda  }{ 1 + \exp{( - \gamma \cdot (e - \tau \cdot E ) )}} + \lambda \\  
    & =   \left( w_y \cdot \frac{1}{ 1 + \exp{( - \gamma \cdot (e - \tau \cdot E ) )}} + 1 \right) \cdot   \lambda  \\
    & =  \left( w_y \cdot \sigma_{\gamma,\tau}(e)  + 1 \right) \cdot   \lambda ,   \\ 
\end{aligned} 
\label{equation_lambda_hat_rewrite}
\end{equation}   
where $\sigma_{\gamma,\tau}$ denotes the proposed variant of sigmoid function, $e$ and $E$ are the current epoch and total number of epochs, respectively. The hyper-parameters $\gamma$ and $\tau$ are used to modulate the steepness and centrality of the sequence as shown in Fig.~\ref{figure_sigmoid_functions_gamma_tau}. $w_y$ is the class weight information, and we quantify it as $w_y = \text{Freq}[y] - 1$, where $\text{Freq}[\cdot]$ refers to the normalized inverse class frequency. Effective number~\cite{DBLP:conf/cvpr/CuiJLSB19} is another common strategy that can describe class information but introduces additional hyper-parameters particularly in many incremental steps. 
 
To further interpret Eq.~\ref{equation_lambda_hat_rewrite}, we view it as a scaling transformation of the coefficient $\lambda$. The  scaling multiplicative factor $(w_y \cdot \sigma_{\gamma,\tau}(e)  + 1 )$ consists of two components: the class weight information component $w_y$ and the non-stationary time series component $\sigma_{\gamma,\tau}(e)$. In this case, $\sigma_{\gamma,\tau}(e)$ assigns a non-stationary sequence process to the model training in CIL, by dynamically adjusting the activation level of weights $w_y$ through outputting a value between 0 and 1 based on the current epoch $e$. Using the normalized inverse class frequency $\text{Freq}[\cdot]$, $w_y$ is utilized to adjust the interpolation coefficient $\lambda$ range in the label space, which expands the range of rare old task samples (where $w_y$ is positive) and contracts range for a large number of new task samples (where $w_y$ is negative). We then have the new mean function $\hat{\mu}(e)$ of $\hat{\lambda}(e, y, \lambda)$,      
  
\begin{equation}
    \hat{\mu}(e) =   \mathbb{E}_{\hat{\lambda}(e)}  =  \frac{  w_y \cdot \sigma_{\gamma,\tau}(e) + 1 }{2} . 
\label{equation_mean_function_hat_lambda}
\end{equation}  
While there are some prior works on dynamic adjustments and weighting~\cite{DBLP:conf/cvpr/ZhouCWC20,DBLP:conf/nips/CaoWGAM19}, our approach  introduces the mixed DA method and models the interpolation coefficients $\lambda$ with non-stationary functions for CIL scenarios. As shown in Fig.~\ref{figure_sigmoid_function_old_weight} and Fig.~\ref{figure_sigmoid_function_new_weight}, we provide two time series process examples corresponding to $\hat{\lambda}$, where one $w_y$ is positive $0.5$ and the other $w_y$ is negative $0.5$. The total number of epochs is $E=240$, $\gamma$ and $\tau$ are set to $0.5$ and $0.6$, respectively. Compared with  Eq.~\ref{equation_mean_function_constant} and Fig.~\ref{figure_sigmoid_function_default_mean}, we introduce not only weight class information but also dynamic non-stationary property for time series processes. Then, the two labels are mixed as    
\begin{equation}
    \begin{aligned} 
    \tilde{y}_{i,j} & = \hat{\lambda}(e, y_i, \lambda) \cdot y_i + \hat{\lambda}(e, y_j, 1-\lambda) \cdot {y_j} . 
    \end{aligned} 
\label{equation_cutmix_label} 
\end{equation}   
After the mixed samples $(\tilde{\mathbf{x}}_{i,j}, \tilde{y}_{i,j})$ are fed into the dynamic architecture model for forward propagation, the cross entropy loss is usually used as the training objective for classification learning.

\subsection{Learning Dynamic Feature with vMF Classifier} 
\label{section_learning_dynamic_features_vmf}

As shown in section~\ref{section_dynamic_architecture} and Eq.~\ref{equation_dynamic_architecture}, the accumulating feature representations of the dynamic architecture may contain a significant amount of knowledge learned from old tasks to mitigate catastrophic forgetting, 
which encourages us to explore learning better features 
from a distribution modeling perspective.    
To this end, we extend the softmax probability $p(y_i | \mathbf{z})$ from Euclidean space to hypersphere surface space~\cite{DBLP:conf/cvpr/LiuWYLRS17,DBLP:conf/iccv/ScottGM21}. Here, we omit the mix operation symbols $i$ and $j$ in previous section for simplicity.     
\begin{equation}
\begin{aligned} 
    p(y_i | \mathbf{z}) 
    & = \frac{ \exp \left( \mathbf{W}_i^\top \mathbf{z} \right) }{ \sum_{j} \exp \left( \mathbf{W}_j^\top \mathbf{z}  \right)  }  \\
    & = \frac{ \exp \left( \| \mathbf{W}_i^\top \|  \|\mathbf{z}\|  \cdot \bar{\mathbf{W}}_i^\top  \bar{\mathbf{z}} \right) }{ \sum_{j} \exp \left(  \| \mathbf{W}_j^\top \|  \|\mathbf{z}\| \cdot  \bar{\mathbf{W}}_j^\top  \bar{\mathbf{z}} \right)  } \\  
    & \approx \frac{ \exp \left( s \cdot \bar{\mathbf{W}}_i^\top  \bar{\mathbf{z}} \right) }{ \sum_{j} \exp \left(  s   \cdot  \bar{\mathbf{W}}_j^\top  \bar{\mathbf{z}} \right)  } \\
    & =  \frac{ \exp \left( s \cdot \langle   
 \bar{\mathbf{W}}_i, \bar{\mathbf{z}} \rangle \right) }{ \sum_{j} \exp \left(  s   \cdot \langle   
 \bar{\mathbf{W}}_j , \bar{\mathbf{z}} \rangle \right)  } , 
\end{aligned}   
\label{equation_probability_e2h} 
\end{equation}
where $\mathbf{z} = \mathcal{F}(\mathbf{x})\in \mathbb{R}^d$ denotes the feature  
corresponding class $y_i$, and $\bar{\mathbf{W}}_i = \frac{\mathbf{W}_i}{\|\mathbf{W}_i\|}$, $\bar{\mathbf{z}} = \frac{\mathbf{z}}{\|\mathbf{z}\|}$. The scaling factor $s$ can be viewed as a pre-fixed hyper-parameter~\cite{DBLP:conf/cvpr/WangWZJGZL018} or as a learnable parameter~\cite{DBLP:conf/eccv/DouillardCORV20}.

\textbf{vMF classifier is a feature distribution model without additional parameter estimation.}   
Based on directional  statistics~\cite{mardia2000directional,sra2018directional}, we can model Eq.~\ref{equation_probability_e2h} with mixture von Mises-Fisher (vMF) distributions~\cite{DBLP:journals/corr/HasnatBMGC17}, and the normalized feature  $\bar{\mathbf{z}}$ lies on the unit hyper-sphere surface $\mathbb{S}^{d-1}$. Then, the definition of the vMF probability density function for class $y_i$ is: 
\begin{equation}  
    p(\bar{\mathbf{z}} |  y_i ) = C_d(\kappa_i) \cdot \exp \left( \kappa_i \cdot \langle \bar{\boldsymbol{\mu}}_i , \bar{\mathbf{z}} \rangle \right) , 
\label{equation_vmf_dist_class_i}
\end{equation} 
where $\kappa_i$ and $\bar{\boldsymbol{\mu}}_i$ are concentration parameter and mean direction for class $y_i$, respectively. The normalization constant $C_d(\kappa_i)$ is equal to
\begin{equation} 
    C_d(\kappa_i) = \frac{\kappa_i^{d/2-1}}{(2 \pi)^{d/2} \cdot I_{d/2-1}(\kappa_i) } , 
\end{equation}
where $I_v$ denotes the modified Bessel function of the first kind at order $v$. Eq.~\ref{equation_vmf_dist_class_i} expresses that the normalized features $\bar{\mathbf{z}}$ of the same class follow to a vMF distribution, which is determined by two parameters, $\kappa$ and $\boldsymbol{\mu}$. These parameters can be further viewed as learnable parameters that can be seamlessly integrated into the final classifier during model training as shown in Eq.~\ref{equation_probability_e2h}. Using Bayes theorem, the posterior probability for predicting class $y_i$ given $\bar{\mathbf{z}}$ is formulated as: 
\begin{equation}
    p(y_i| \bar{\mathbf{z}} ) = \frac{ p(\bar{\mathbf{z}} | y_i  ) p(y_i)  }{\sum_j  p(\bar{\mathbf{z}}  | y_j ) p(y_j) } = \frac{ \exp \left( \kappa \cdot \langle \bar{\boldsymbol{\mu}}_i , \bar{\mathbf{z}} \rangle \right) }{ \sum_j \exp \left( \kappa \cdot \langle \bar{\boldsymbol{\mu}}_j , \bar{\mathbf{z}} \rangle \right) } ,
\label{equation_posterior_probability_vmf}
\end{equation}
where we assume that the concentrations of different classes are the same in mixture vMF distributions, $\{\kappa_i\}_{i=1}^{C} = \kappa$, and the class prior have been taken into account in our proposed memory-centric mix as discussed in section~\ref{section_memory_centric_mix}. Therefore, we established the vMF  classifier to model the accumulated features $\bar{\mathbf{z}}$ in Eq.~\ref{equation_posterior_probability_vmf}, where $\bar{\boldsymbol{\mu}}_i =  \frac{\mathbf{W}_i} {\|\mathbf{W}_i\|}$ is the normalized weight of the classifier, and $\kappa$ is a learnable parameter similar to $s$ in Eq.~\ref{equation_probability_e2h}. Moreover, the dynamic features follow the known mixture vMF distributions, which provides an analytical model for further investigation, leading to the derivation of our proposed feature matching loss as discussed in section~\ref{section_feature_matching}.

\textbf{vMF classifier is a feature learning model.} As stated at the beginning of this section, we aim to learn a good feature representation in dynamic architecture. We first rewrite the  Eq.~\ref{equation_dynamic_architecture} as: 
\begin{equation}
    \mathbf{z} = \mathcal{F}_t(\mathbf{x}) = [ f_t(\mathbf{x}) ; \mathcal{F}_{t-1}(\mathbf{x}) ] = [\mathbf{z}^{(\mathrm{n})};\mathbf{z}^{(\mathrm{o})}],    
\label{equation_dynamic_architecture_with_z} 
\end{equation}
where $\mathbf{z}^{(\mathrm{n})}$and $\mathbf{z}^{(\mathrm{o})}$ denote the new trainable features and old frozen features, respectively. Similarly, we employ the the negative log-likelihood transformation of Eq.~\ref{equation_posterior_probability_vmf} as:     
\begin{equation}     
\begin{aligned} 
\mathcal{L}_\mathrm{NLL} & =   -  \log  \frac{ \exp \left( \kappa \cdot \langle \bar{\boldsymbol{\mu}}_i , \bar{\mathbf{z}}    \rangle \right) }{ \sum_j \exp \left( \kappa \cdot \langle \bar{\boldsymbol{\mu}}_j , \bar{\mathbf{z}}  \rangle \right) }  \\   
& = \log \Biggr[  1 + \sum^C_{j \not= i} \exp \left( \kappa \cdot ( \langle \bar{\boldsymbol{\mu}}_j , \bar{\mathbf{z}} \rangle   - \langle \bar{\boldsymbol{\mu}}_i , \bar{\mathbf{z}}   \rangle ) \right) \Biggr]  .  
\label{equation_nll_softmax}
\end{aligned} 
\end{equation}   
With the vMF distribution modeling, the mean direction can be obtained as the normalized arithmetic mean of the feature samples, $\bar{\boldsymbol{\mu}}_i = \frac{\boldsymbol{\mu}_i}{ \| \boldsymbol{\mu}_i \|}$, where $\boldsymbol{\mu}_i = \frac{1}{N}  \sum_{n=1}^N \bar{\mathbf{z}}$ and $\bar{\mathbf{z}} \sim p(\bar{\mathbf{z}}|y_i)$.  The key component of Eq.~\ref{equation_nll_softmax} can be rewritten as 
\begin{equation}
\begin{aligned}
    \kappa \cdot \langle \bar{\boldsymbol{\mu}} , \bar{\mathbf{z}}    \rangle & = \kappa   \cdot  \left( \langle \bar{\boldsymbol{\mu}}^{(\mathrm{n})}   , \bar{\mathbf{z}}^{(\mathrm{n})}  \rangle +  \langle \bar{\boldsymbol{\mu}}^{(\mathrm{o})}   ,  \bar{\mathbf{z}}^{(\mathrm{o})}   \rangle \right) \\
 & = \kappa   \cdot  \left( \langle \frac{1}{N}    \sum_{n=1}^{N}      {\bar{\mathbf{z}}_{n}^{(\mathrm{n})}}  ,   \bar{\mathbf{z}}^{(\mathrm{n})}  \rangle   +  \langle \frac{1}{N}    \sum_{n=1}^{N}      {\bar{\mathbf{z}}_{n}^{(\mathrm{o})}} ,   \bar{\mathbf{z}}^{(\mathrm{o})}    \rangle \right) \\ 
 & = \frac{1}{N}    \sum_{n=1}^{N}    \kappa   \cdot  \langle    {\bar{\mathbf{z}}_{n}^{(\mathrm{n})}}  ,   \bar{\mathbf{z}}^{(\mathrm{n})}  \rangle + \mathrm{const} , 
\end{aligned}  
\label{equation_key_component}
\end{equation}   
where $\bar{\boldsymbol{\mu}}$ is also split into two parts $\bar{\boldsymbol{\mu}}^{(\mathrm{n})}$ and $\bar{\boldsymbol{\mu}}^{(\mathrm{o})}$.  $\bar{\mathbf{z}}_{n}^{(\mathrm{n})}$ denotes the other new features belonging to the same class as $\bar{\mathbf{z}}^{(\mathrm{n})}$. Here we omit the normalization scaling operation in arithmetic mean function. Since the old feature extractor is frozen, we replace the result of the old features with a constant. As the $\mathrm{LogSumExp}$ function is a smooth  approximation of the $\mathrm{max}$ function~\cite{boyd2004convex}, we can rewrite  Eq.~\ref{equation_nll_softmax} as             
\begin{equation}     
\begin{aligned} 
\mathcal{L}_\mathrm{NLL} 
& \approx \max{ 
\{0\} \cup   \left\{ \kappa \cdot ( \langle \bar{\boldsymbol{\mu}}_j , \bar{\mathbf{z}} \rangle   - \langle \bar{\boldsymbol{\mu}}_i , \bar{\mathbf{z}}   \rangle )  \right\}_{j \not= i}^C 
} . \\  
\end{aligned}
\label{equation_nll_approx_max}
\end{equation} 
Then, we can further substitute Eq.~\ref{equation_key_component} into the second part of Eq.~\ref{equation_nll_approx_max} to get         
\begin{equation}
    \left\{ \kappa \cdot   \left( \frac{1}{N_j}   \sum_{n=1}^{N_j}  \langle    {\bar{\mathbf{z}}_{n,-}^{(\mathrm{n})}} ,   \bar{\mathbf{z}}^{(\mathrm{n})}  \rangle -    \frac{1}{N_i} \sum_{n=1}^{N_i} \langle     {\bar{\mathbf{z}}_{n,+}^{(\mathrm{n})}} ,     \bar{\mathbf{z}}^{(\mathrm{n})}  \rangle   \right)   \right\}_{j \not= i}^C ,
\label{equation_max_second_part_set}
\end{equation}  
where we represent the samples belonging to the same class $i$ and different classes $j, j\neq i$ as positive samples $+$ and negative samples $-$, respectively. Compared with triplet  loss~\cite{DBLP:conf/cvpr/SchroffKP15},  
\begin{equation}
    \mathcal{L}_{\mathrm{Triplet}} = \max{ \{0, \langle    {\bar{\mathbf{z}}_{-}^{(\mathrm{n})}} ,      \bar{\mathbf{z}}^{(\mathrm{n})}  \rangle-  \langle    {\bar{\mathbf{z}}_{+}^{(\mathrm{n})}} ,     \bar{\mathbf{z}}^{(\mathrm{n})}  \rangle \}} ,    
\label{equation_triplet_loss}
\end{equation}  
we observe that the loss function of the vMF classifier shares a strikingly similar form with that of the triplet loss. Different from unified loss perspective~\cite{DBLP:conf/cvpr/SunCZZZWW20} and vMF based N-paired loss learning~\cite{DBLP:journals/pr/ZheCY19}, this paper explores the dynamic new  features in CIL from the perspective of triplet loss with vMF modeling.     
(\romannumeral1) Multiple positive samples. Unlike triplet loss, which randomly selects only one positive sample, Eq.~\ref{equation_max_second_part_set} takes into account $N_i$ positive samples. (\romannumeral2) Multiple negative samples. Similar to the above, it uses $N_j$ negative samples. By constraining the anchor based on the average direction of positive and negative samples formed by multiple groups of samples, Eq.~\ref{equation_max_second_part_set} can make better use of the samples and reduce variance for feature learning.   (\romannumeral3) Multi-class max function. By using the maximum similarity value among negative samples from different classes as the loss value, Eq.~\ref{equation_max_second_part_set} can identify the most similar group of negative samples from a class and penalize them. In the extreme case where there is only one additional class with a single sample in each class, the loss function simplifies into the form of triplet loss.

\textbf{Brief summary.} Our findings suggest that the vMF classifier can naturally induce the distribution probability density function of the global features $\bar{\mathbf{z}}  = [\bar{\mathbf{z}}^{(\mathrm{n})};\bar{\mathbf{z}}^{(\mathrm{o})}]$, and performing  discriminative classification based on these features to enable the learning of new task knowledge while mitigating the forgetting of old task knowledge. Based on the vMF distribution modeling, we can further derive our proposed dynamic feature matching loss function as discussed in the next section. Moreover, by rewriting the loss function of the vMF classifier, we can find that it implicitly imposes constraints on the newly learned dynamic features $\bar{\mathbf{z}}^{(\mathrm{n})}$, i.e., separates features of different classes and brings features of the same class closer.

\subsection{Dynamic Feature Matching} 
\label{section_feature_matching}

The parameters $\kappa$ and $\bar{\boldsymbol{\mu}}$, estimated directly by gradient descent, play a very important role in the probability output in the final classifier. On the other hand, with vMF modeling in section~\ref{section_learning_dynamic_features_vmf}, the $\kappa$ and $\bar{\boldsymbol{\mu}}$ describe the properties of the dynamic features obtained by the backbone extractors. However, the performance is still limited by the issue of dynamic features and vMF classifier mismatch. Empirical results indicate a discernible gap in accuracy between CNN and NME as shown in Tab.~\ref{table_main_ablation_comparision}.

Therefor, we propose the online batch update dynamic feature matching loss to alleviate this issue. Specifically, we compute the class mean for each batch of samples, and consider it to fit a new vMF distribution.
\begin{equation} 
    \tilde{\boldsymbol{\mu}}_i = \frac{{\boldsymbol{\mu}}_i}{\|{\boldsymbol{\mu}}_i\|}, \quad   {\boldsymbol{\mu}}_i =  \frac{1}{B_i} \sum_{n=1}^{B_i}   \bar{\mathbf{z}}_n   ,  
\end{equation}
where $\{\bar{\mathbf{z}}_n\}_{n=1}^{B_i}$ are the latent dynamic features belonging to the class $y_i$ in the mini-batch, which are sampled from $\bar{\mathbf{z}}_n \sim p(\bar{\mathbf{z}} |y_i)$ in Eq.~\ref{equation_vmf_dist_class_i}. We compute the class mean ${\boldsymbol{\mu}}_i$, then calculate the normalized class direction $\tilde{\boldsymbol{\mu}}_i$, and $B_i$ denote the the number of features of class $y_i$ in the mini-batch. Now, we model the new mixture vMF distributions as   
\begin{equation}  
    \tilde{p}(\bar{\mathbf{z}} |  y_i ) = C_d(\kappa_i) \cdot \exp \left( \kappa_i \cdot \langle \tilde{\boldsymbol{\mu}}_i , \bar{\mathbf{z}} \rangle \right) ,
\label{equation_new_vmf_dist_class_i}
\end{equation} 
where $\{\kappa_i\}_{i=1}^{C} = \kappa$ are set to the same parameters as in Eq.~\ref{equation_posterior_probability_vmf}. To quantify the mismatch problem, we simply utilize KL divergence to analytically measure the mismatch between Eq.~\ref{equation_vmf_dist_class_i} and Eq.~\ref{equation_new_vmf_dist_class_i} as
\begin{equation}
\begin{aligned}
    D_\mathrm{KL} (p(\bar{\mathbf{z}} |  y_i ) \| \tilde{p}(\bar{\mathbf{z}} |  y_i ) ) & = \int  p(\bar{\mathbf{z}} |  y_i )  \cdot \log \frac{p(\bar{\mathbf{z}} |  y_i )}{\tilde{p}(\bar{\mathbf{z}} |  y_i )}  \mathrm{d}   \bar{\mathbf{z}}  \\
    & = \kappa \cdot A_d(\kappa)  \cdot (1 - \langle  \bar{\boldsymbol{\mu}}_i , \tilde{\boldsymbol{\mu}}_i \rangle ) ,
\end{aligned}
\label{equation_kl_divergence}
\end{equation}
where $A_d(\kappa) = I_{d/2}(\kappa) / I_{d/2-1}(\kappa)$.

\begin{proof}
According to Eq.~\ref{equation_kl_divergence}, we have   
    \begin{equation}
    \begin{aligned}
        D_\mathrm{KL} 
        & = \int  p(\bar{\mathbf{z}} |  y_i )  \cdot \log \frac{p(\bar{\mathbf{z}} |  y_i )}{\tilde{p}(\bar{\mathbf{z}} |  y_i )}  \mathrm{d}   \bar{\mathbf{z}} \\ 
        & = \int  p(\bar{\mathbf{z}} |  y_i )  \cdot \left[ \log p(\bar{\mathbf{z}} |  y_i ) - \log \tilde{p}(\bar{\mathbf{z}} |  y_i ) \right]   \mathrm{d}   \bar{\mathbf{z}} \\
        & = \int   p(\bar{\mathbf{z}} |  y_i )  \cdot \left[ \log \frac{C_d(\kappa)}{C_d(\kappa)} + \log \frac{ \exp ( \kappa \cdot \langle  \bar{\boldsymbol{\mu}}_i , \bar{\mathbf{z}} \rangle ) }{ \exp  ( \kappa \cdot \langle \tilde{\boldsymbol{\mu}}_i , \bar{\mathbf{z}} \rangle )  } \right] \mathrm{d}   \bar{\mathbf{z}} \\
        & = \int  p(\bar{\mathbf{z}} |  y_i )  \cdot \left[ \log \frac{C_d(\kappa)}{C_d(\kappa)}  +   \kappa \cdot ( \langle \bar{\boldsymbol{\mu}}_i , \bar{\mathbf{z}} \rangle  - \langle \tilde{\boldsymbol{\mu}}_i , \bar{\mathbf{z}} \rangle  )    \right] \mathrm{d}   \bar{\mathbf{z}} \\ 
        & = \kappa \cdot \int   p(\bar{\mathbf{z}} |  y_i )   \cdot   \left( \bar{\boldsymbol{\mu}}_i^\top - \tilde{\boldsymbol{\mu}}_i^\top  \right) \cdot  \bar{\mathbf{z}} \,  \mathrm{d}   \bar{\mathbf{z}}  \\
        & = \kappa \cdot \left( \bar{\boldsymbol{\mu}}_i^\top - \tilde{\boldsymbol{\mu}}_i^\top  \right) \cdot  \int    \bar{\mathbf{z}} \cdot  p(\bar{\mathbf{z}} |  y_i )    \,  \mathrm{d}   \bar{\mathbf{z}}   \\
        & = \kappa \cdot \left( \bar{\boldsymbol{\mu}}_i^\top - \tilde{\boldsymbol{\mu}}_i^\top  \right)  \cdot \mathbb{E}_{\bar{\mathbf{z}} \sim p(\bar{\mathbf{z}} |  y_i )  }   [ \bar{\mathbf{z}}  ] .  
    \end{aligned}  
    \end{equation}  
Inspired  by~\cite{DBLP:conf/eccv/WangFHFLH22,diethe2015note}, for the vMF distribution $p(\bar{\mathbf{z}} |  y_i )$ with mean direction $\bar{\boldsymbol{\mu}}_i$ and concentration $\kappa$, the expected value is $A_d(\kappa) \cdot \bar{\boldsymbol{\mu}}_i$, where $A_d(\kappa) = I_{d/2}(\kappa) / I_{d/2-1}(\kappa)$. Then, we have 
\begin{equation}
\begin{aligned}  
    D_\mathrm{KL} & = \kappa \cdot \left( \bar{\boldsymbol{\mu}}_i^\top - \tilde{\boldsymbol{\mu}}_i^\top  \right)  \cdot A_d(\kappa) \cdot  \bar{\boldsymbol{\mu}}_i  \\
    & =  \kappa \cdot A_d(\kappa)  \cdot (1 - \langle  \bar{\boldsymbol{\mu}}_i , \tilde{\boldsymbol{\mu}}_i \rangle ) . 
\end{aligned}
\end{equation}
\end{proof}

Thus, our proposed dynamic feature matching loss function is defined as  
\begin{equation} 
    \mathcal{L}_\mathrm{Ma} = \kappa \cdot A_d(\kappa)  \cdot (1 - \langle  \bar{\boldsymbol{\mu}}_i , \tilde{\boldsymbol{\mu}}_i \rangle ) ,
\label{equation_loss_functions_feature_match}
\end{equation}
where $\kappa$ and $\bar{\boldsymbol{\mu}}_i$ are parameters of vMF classifier. $\tilde{\boldsymbol{\mu}}_i$ denotes the normalized class mean feature vector in the mini-batch. The proposed matching loss aligns the dynamic feature with the vMF classifier, leading to improved performance.


\subsection{Overview}

In this subsection, we provide an overview of our proposed method. Fig.~\ref{fig_overview} illustrates the overall framework, depicting both the workflow and the loss functions for model training.

The training dataset consists of the current task data and a small amount of past task data $\mathcal{D}_t \cup \mathcal{M}_t$. Based on the dataset, we first employ the memory-centric mix to construct mixed samples and labels $(\tilde{\mathbf{x}}_{i,j}, \tilde{y}_{i,j})$ ($(\mathbf{x}, y)$ for simplicity), as shown in Eq.~\ref{equation_cutmix_samples}, Eq.~\ref{equation_lambda_hat_rewrite}, and Eq.~\ref{equation_cutmix_label}. Fig.~\ref{fig_overview} presents our MC-Mix strategy, which encompasses non-stationary sequences involving a sigmoid function, along with class weight information regarding the sample counts. Next, the mixed data is fed into a dynamic architecture, which consists of frozen previous feature extractors and a trainable new feature extractor, to obtain dynamic features $\mathbf{z} = [\mathbf{z}^{(\mathrm{n})};\mathbf{z}^{(\mathrm{o})}]$ as shown in Eq.~\ref{equation_dynamic_architecture}. Finally, a vMF classifier is employed for the final probability prediction output in Eq.~\ref{equation_posterior_probability_vmf}.

The training loss functions are composed of three key components: the negative log-likelihood (NLL) loss, the auxiliary classifier (Aux) loss, and the dynamic feature matching (Ma) loss, as shown in Fig.~\ref{fig_overview}. The NLL loss  $\mathcal{L}_\mathrm{NLL}$ in Eq.~\ref{equation_nll_softmax} shares properties with the commonly used softmax cross entropy loss in Eq.~\ref{equation_cross_entropy_default_loss_function} for classification learning,   
implicitly separating dissimilar instances and bringing similar instances closer under vMF modeling, as described in section~\ref{section_learning_dynamic_features_vmf}.  The $\mathcal{L}_\mathrm{Aux}$ loss function in Eq.~\ref{equation_aux_loss_function} is employed to distinguish between data from the old and new tasks. The $\mathcal{L}_\mathrm{Ma}$ loss in Eq.~\ref{equation_loss_functions_feature_match} is used for dynamic feature matching. The total training loss is as follows    
\begin{equation}
    \mathcal{L} = \mathcal{L}_\mathrm{NLL} + \eta_\mathrm{Aux} \cdot \mathcal{L}_\mathrm{Aux} + \eta_\mathrm{Ma} \cdot  \mathcal{L}_\mathrm{Ma}, 
\end{equation}
where $\eta_\mathrm{Aux}$ and $\eta_\mathrm{Ma}$ are hyper-parameters to control the effect of the loss functions. 

\section{Experiment} 

\begin{table*}
\caption{Performance comparison on the CIFAR-100 benchmarks. Average accuracy ($\%$) over all steps are reported. We run experiments using three different class orders and report their mean and standard deviation.} 
\begin{center}  
\begin{tabular}{l|cccccc}
\toprule
Dataset & \multicolumn{6}{c}{CIFAR-100} \\ 
\midrule      
Methods & $\,\,\,$ B0 5 steps $\,\,\,$ & $\,\,\,$ B0 10  steps $\,\,\,$ & $\,\,\,$ B0 20 steps $\,\,\,$ & $\,\,\,$ B50 5 steps $\,\,\,$ & $\,\,\,$ B50 10 steps $\,\,\,$ & $\,\,\,$ B50 25 steps $\,\,\,$ \\   
\midrule 
iCaRL~\cite{DBLP:conf/cvpr/RebuffiKSL17}  &  $67.52_{\pm 1.45}$ & $64.06_{\pm 1.41}$ & $62.30_{\pm 1.48}$   &  $57.51_{\pm 0.87}$ & $53.86_{\pm 0.99}$ & $50.39_{\pm 0.81}$  \\
WA~\cite{DBLP:conf/cvpr/ZhaoXGZX20} & $69.15_{\pm 1.00}$ & $67.21_{\pm 1.34}$ & $65.31_{\pm 1.56}$   & $62.90_{\pm 0.92}$ & $57.97_{\pm 0.58}$ & $51.90_{\pm 1.09}$ \\
PODNet~\cite{DBLP:conf/eccv/DouillardCORV20} &  $64.11_{\pm 1.19}$ & $54.60_{\pm 1.21}$ & $46.82_{\pm 1.09}$ & $64.98_{\pm 0.41}$ & $63.16_{\pm 0.80}$ & $60.79_{\pm 0.92}$ \\        
AFC~\cite{DBLP:conf/cvpr/class2022kang} & $66.41_{\pm 1.10}$ & $60.40_{\pm 1.51}$ & $55.19_{\pm 1.58}$ & $65.84_{\pm 0.57}$ & $64.72_{\pm 0.60}$ & $63.79_{\pm 0.64}$ \\ 
DER~\cite{DBLP:conf/cvpr/YanX021} & $73.14_{\pm 0.86}$ & $71.52_{\pm 0.83}$ & $70.32_{\pm 1.43}$  & $68.97_{\pm 0.45}$ & $67.51_{\pm 0.36}$  & $64.89_{\pm 0.53}$ \\    
$\text{FOSTER}^\dagger$~\cite{DBLP:conf/eccv/WangZYZ22} & $75.98_{\pm 0.96}$ & $73.98_{\pm 1.16}$ & $71.54_{\pm 1.71}$ & $72.62_{\pm 0.79}$  & $69.59_{\pm 0.79}$ & $66.17_{\pm 0.46}$ \\   
\midrule 
Ours & $74.66_{\pm 0.86}$ & $74.03_{\pm 1.18}$ & $72.93_{\pm 1.33}$  & $70.73_{\pm 0.75}$ & $69.85_{\pm 0.43}$ & $67.89_{\pm 0.53}$ \\
$\text{Ours}^\dagger \, \textit{w}/\mathrm{Aug}$  $\,\,\,\,\,$ & $76.37_{\pm 0.78}$ & $76.26_{\pm 1.01}$ & $75.31_{\pm 0.90}$ & $73.32_{\pm 0.62}$ & $72.58_{\pm 0.44}$ & $71.62_{\pm 0.66}$ \\ 
\bottomrule 
\end{tabular} 
\end{center}
\label{table_cifar100_benchmakrs}
\end{table*}

\begin{figure*}[t]
\centering    
\includegraphics[width=0.99\textwidth]{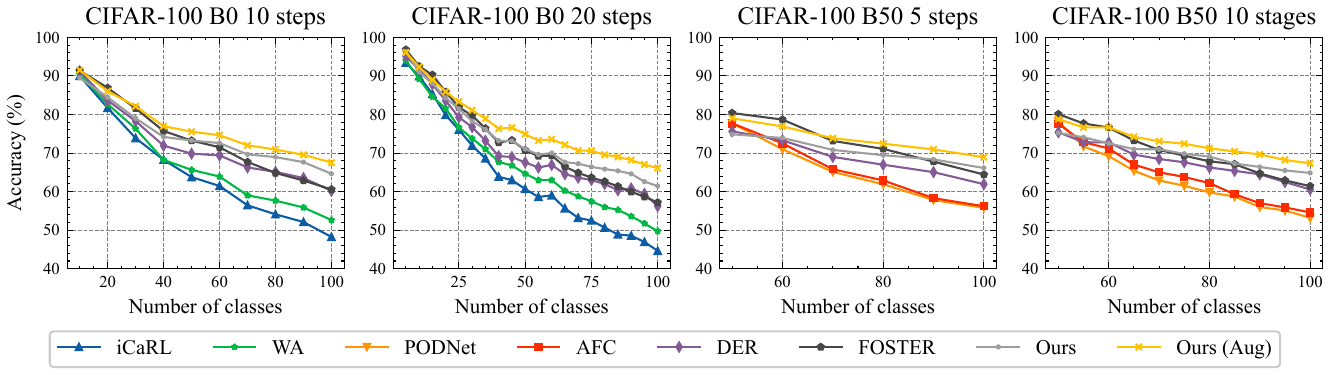} 
\caption{Performance comparison for each step. From left to right: CIFAR-100 B0 10 steps, CIFAR-100 B0 20 steps, CIFAR-100 B50 5 steps, and CIFAR-100 B50 10 steps. }
\label{fig_plot_compare_accuracy_cifar100} 
\end{figure*}   
 

\subsection{Datasets and Benchmarks}

We employ CIFAR-100~\cite{Krizhevsky_2009_17719}, ImageNet-100, and ImageNet-1000~\cite{DBLP:conf/cvpr/DengDSLL009} for evaluation. CIFAR-100 is a dataset of $32 \times 32$ RGB color images, consisting of 50,000 training images and 10,000 testing images over 100 classes. ImageNet-1000 is a large dataset commonly used ILSVRC, which consists of 1000 classes with about 1.28M training images and 50K validation images. ImageNet-100 is a subset of ImageNet-1000, where we select the same 100 classes based on 
~\cite{DBLP:conf/eccv/DouillardCORV20}.

For the test protocols, we construct a wide range of diverse settings based on the above three datasets. (\romannumeral1) \textbf{CIFAR-100:} A uniform backbone network configuration of ResNet-32~\cite{DBLP:conf/cvpr/RebuffiKSL17,DBLP:conf/cvpr/HeZRS16} is employed for fair comparisons.  Firstly, we adopt the testing protocol described in~\cite{DBLP:conf/cvpr/RebuffiKSL17}, where the 100 classes are directly divided into 5, 10, and 20 tasks, with each task involving the incremental learning of 20, 10, and 5 new classes, respectively. These settings are usually refer to as "B0 $T$ steps", where $T$ denotes the total number of tasks according to~\cite{DBLP:conf/cvpr/YanX021}. The memory size is fixed to a total of 2,000 exemplars during incremental steps. On the other hand, another common used testing protocol~\cite{DBLP:conf/eccv/DouillardCORV20} involves training half of the classes, namely 50 classes, in the first task, followed by incremental learning of new classes in subsequent tasks. The remaining 50 classes are divided into 5, 10 and 25 tasks, where each task requires incremental learning of 10, 5 and 2 new classes, respectively. This protocol is referred to as "B50 $T$ steps", where $T$ denotes the number of the following incremental tasks. And the memory size is fixed at 20 samples per class. (\romannumeral2) \textbf{ImageNet-100:} Similarly, CIFAR-100 and ImageNet-100 have the same number of classes, the division of the incremental tasks and the memory size are set up in the same way. The backbone network is set as ResNet-18~\cite{DBLP:conf/cvpr/HeZRS16} followed by~\cite{DBLP:conf/cvpr/RebuffiKSL17,DBLP:conf/eccv/DouillardCORV20}. (\romannumeral3) \textbf{ImageNet-1000:} The ResNet-18 network backbone is also adopted in ImageNet-1000 dataset. We evaluate our model on B0 10 steps, where each task involves learning 100 classes using 20,000 memory exemplars.   
Moreover, we report the average accuracy~\cite{DBLP:conf/cvpr/RebuffiKSL17} across all tasks as a quantitative metric for all settings. 
Finally, we build a comprehensive and detailed comparison benchmarks that includes a total of 13 different settings.

\subsection{Implementation Details}

We implement our model with PyTorch~\cite{DBLP:conf/nips/PaszkeGMLBCKLGA19}. During model training, we employ the SGD optimizer with a weight decay of 2e-4 across all datasets. Additionally, a linear warm-up strategy is adopted for the initial 10 epochs. For CIFAR-100 benchmarks, we conduct training for a total of 240 epochs using a batch size of 128. The learning rate starts at 0.1 and decays at 100, 170, and 210 epochs with a decay rate of 0.1. The ImageNet-100 benchmarks utilize a learning rate of 0.1 that decays at 80, 130, and 170 epochs by a factor of 0.1, with a total of 200 epochs and a batch size of 128. The ImageNet-1000 benchmark exclusively utilizes MC-Mix as the model, training it for 240 epochs with a batch size of 256. The learning rate also starts at 0.1 and decays at 100, 160, and 200 epochs with a decay rate of 0.1. Moreover, herding exemplars  selection~\cite{DBLP:conf/cvpr/RebuffiKSL17} and weight alignment~\cite{DBLP:conf/cvpr/ZhaoXGZX20} strategies are adopted in our model. 
For MC-Mix, it is only added in the incremental steps after the first task.
We also consider additional improvement,  denoted by  "$\dagger \,  \textit{w}/\mathrm{Aug}$", which adds AutoAugment~\cite{DBLP:conf/cvpr/CubukZMVL19} and color jitter DA methods following the~\cite{DBLP:conf/eccv/WangZYZ22}, and includes the vanilla CutMix in the first task.

\begin{table*}
\caption{Performance comparison on the ImageNet-100 and ImageNet-1000 benchmarks. } 
\begin{center}
\begin{tabular}{l|cccccc|c} 
\toprule 
Dataset & \multicolumn{6}{c|}{ImageNet-100} &   ImageNet-1000\\ 
\midrule 
Methods  & B0 5 steps & B0 10  steps & B0 20 steps & B50 5 steps & B50 10 steps & B50 25 steps & B0 10 steps  \\ 
\midrule  
iCaRL~\cite{DBLP:conf/cvpr/RebuffiKSL17} & $73.85$ & $68.25$ & $64.96$  & - & -  & - &   $38.40$ \\     
WA~\cite{DBLP:conf/cvpr/ZhaoXGZX20} & $75.62$ & $70.16$  &  $66.45$ & - & - & - & $65.67$ \\ 
PODNet~\cite{DBLP:conf/eccv/DouillardCORV20} & - & - & - & $75.54$ & $74.33$ & $68.31$ & - \\   
AFC~\cite{DBLP:conf/cvpr/class2022kang} & - & - & - &  $76.87$ & $75.75$ & $73.34$ & -  \\ 
DER~\cite{DBLP:conf/cvpr/YanX021} & $78.82$ & $77.23$ & $75.18$ & $79.53$ & $78.25$ & $75.17$ & $68.84$ \\ 
$\text{FOSTER}^\dagger$~\cite{DBLP:conf/eccv/WangZYZ22} & $79.04$ & $78.10$ & $74.03$ & $79.74$ &  $77.55$ & $73.17$ & $68.34$ \\  
\midrule 
Ours & $79.75$ & $79.01$ & $77.98$ & $79.56$ & $78.98$ & $76.63$ & $70.52$ \\  
$\text{Ours}^\dagger \, \textit{w}/\mathrm{Aug}$ & $81.35$ & $80.48$ & $80.01$ & $82.25$ & $81.98$ & $78.70$ & $71.01$ \\  
\bottomrule
\end{tabular}
\end{center}
\label{table_imagenet_benchmark}
\end{table*} 

\begin{figure*}[t]
\centering 
\includegraphics[width=0.99\textwidth]{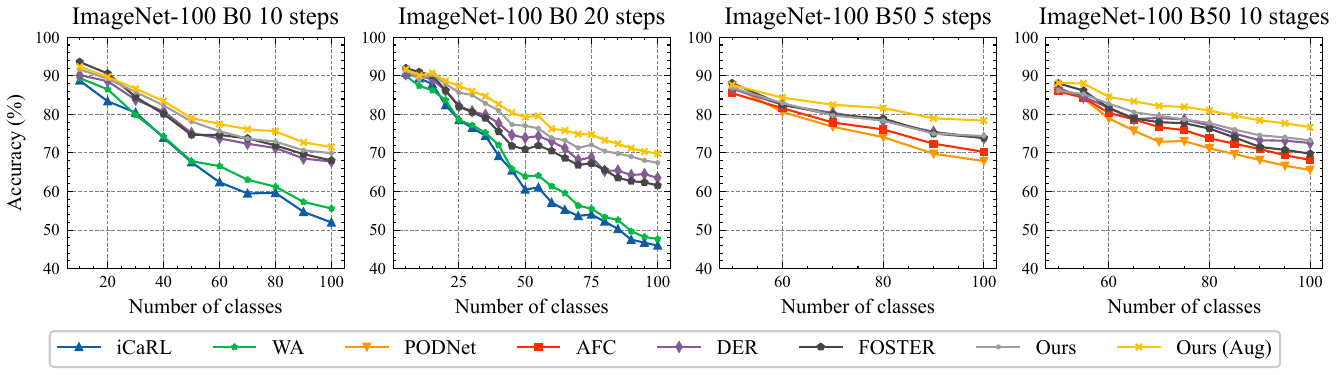} 
\caption{Performance comparison for each step. From left to right: ImageNet-100 B0 10 steps, ImageNet-100 B0 20 steps, ImageNet-100 B50 5 steps, and ImageNet-100 B50 10 steps. }
\label{fig_plot_compare_accuracy_imagenet100}  
\end{figure*}

\begin{table*}
\caption{Ablation Comparison of Components on the CIFAR-100 Benchmarks.}  
\label{table_main_ablation_comparision} 
\begin{center}
\begin{tabular}{ccc|cccc|cccc} 
\toprule
\multicolumn{3}{c|}{Dataset} & \multicolumn{8}{c}{CIFAR-100} \\ 
\midrule 
\multicolumn{3}{c|}{Components} & \multicolumn{4}{c|}{B0 10 steps} & \multicolumn{4}{c}{B50 5 steps}  \\  
\midrule  
Mix & vMF & Ma. & \multicolumn{2}{c}{CNN} & \multicolumn{2}{c|}{NME} & \multicolumn{2}{c}{CNN} & \multicolumn{2}{c}{NME}  \\  
\midrule 
- & - & - & $71.67$ & - & $71.44$ & - & $67.79$ & - & $68.52$ & -  \\  
 $\checkmark$ &&& $74.86$ & $+3.19$  & $72.24$ & $+0.80$ & $67.86$ & $+0.08$ & $68.69$ & $+0.17$ \\ 
 $\checkmark$ & $\checkmark$ && $74.90$ & $+3.23$ & $73.69$ & $+2.25$ & $69.19$ & $+1.40$ & $69.50$ & $+0.98$ \\  
 $\checkmark$ & $\checkmark$ & $\checkmark$ & $\mathbf{74.92}$ & $\mathbf{+3.25}$ & $\mathbf{74.50}$ & $\mathbf{+3.06}$ & $\mathbf{70.90}$ & $\mathbf{+3.21}$ & $\mathbf{70.72}$ & $\mathbf{+2.20}$  \\  
\bottomrule 
\end{tabular}
\end{center}
\end{table*}

\subsection{Main Results}

We compare our method with various models, including iCaRL~\cite{DBLP:conf/cvpr/RebuffiKSL17}, WA~\cite{DBLP:conf/cvpr/ZhaoXGZX20}, PODNet~\cite{DBLP:conf/eccv/DouillardCORV20}, AFC~\cite{DBLP:conf/cvpr/class2022kang}, DER~\cite{DBLP:conf/cvpr/YanX021}, and FOSTER~\cite{DBLP:conf/eccv/WangZYZ22}. We reproduce the results of iCaRL, WA, and DER based on the PyCIL framework~\cite{DBLP:journals/corr/abs-2112-12533}. For PODNet, AFC, and FOSTER, we use the official code for reproduction. FOSTER additionally use "$\dagger$" as a marker to indicate the adaption of strong DA. The ImageNet-1000 benchmark results are directly reported from~\cite{DBLP:conf/cvpr/YanX021} and~\cite{DBLP:conf/eccv/WangZYZ22}.

For CIFAR-100 dataset, Table~\ref{table_cifar100_benchmakrs} and Fig.~\ref{fig_plot_compare_accuracy_cifar100}  summarizes the quantitative results of different methods. We run experiments using three different class orders and report their averages and standard deviations. 
Overall, our model consistently outperforms all previous methods on both B0 and B50 benchmarks, particularly when augmented with strong DA techniques ("$\dagger \,  \textit{w}/\mathrm{Aug}$"), demonstrating superior performance on the CIFAR-100 dataset. The DER and FOSTER, as dynamic architecture based models, demonstrate significant performance improvements over previous KD based methods, validating the promising prospects of this direction. Specifically, under the protocol setting of B0 10 steps, we improve the average accuracy from $71.52 \%$ to $74.03 \%$ ($+2.51 \%$). When further applying strong DA, our performance increase from $73.98 \%$ to $76.26 \%$ ($+2.28 \%$), which validate the effectiveness of our proposed model. In an alternative common setting, namely the B50 10 steps benchmark, our method demonstrates a commendable average accuracy of $72.98 \%$. It is noteworthy that the integration of the strong DA techniques with our model consistently lead to further improvements (e.g. from $69.85 \%$ to $72.58 \%$ on B50 10 steps). This empirical observation not only substantiates the strong coupling between DA and CIL, presenting a promising direction to mitigate catastrophic forgetting by focusing on the quality of training samples, but also serves as validation for the further improvement of our model by the incorporation of strong DA techniques. Meanwhile, as illustrated in Fig.~\ref{fig_plot_compare_accuracy_cifar100}, it is evident that our approach consistently outperforms other methods in nearly every incremental step, especially in the last step where a significant increase in performance is observed and further proves the success of our model.  

For the ImageNet-100 and ImageNet-1000 benchmarks, we achieve similar results to those obtained on the CIFAR-100 dataset, surpassing previous methods across all testing protocols, as shown in Table~\ref{table_imagenet_benchmark} and Fig.~\ref{fig_plot_compare_accuracy_imagenet100}. Specifically, we achieve impressive performance with average accuracies of $80.48 \%$ and $81.98 \%$ on the ImageNet-100 B0 10 and B50 10 steps benchmarks, respectively. Fig.~\ref{fig_plot_compare_accuracy_imagenet100} also showcases the outstanding performance of our model at each incremental step, especially in the last step. For ImageNet-1000 B0 10 steps, we directly adapot the MC-Mix as our model according to the ablation study, and achieve the average accuracies of $70.52 \%$ and $71.01 \%$, which validates the effectiveness of our method. It is worth noting that the performance improvement of the model after the addition of strong DA is not as significant as observed in other benchmarks. We speculate that this could be attributed to the rich quality of samples in the large-scale dataset.

\begin{figure}[t]
\begin{center}
    \subfigure[Average accuracy ($\%$). ]{
    \begin{minipage}[t]{0.263\textwidth}   
    \centering     \includegraphics[width=1\textwidth] {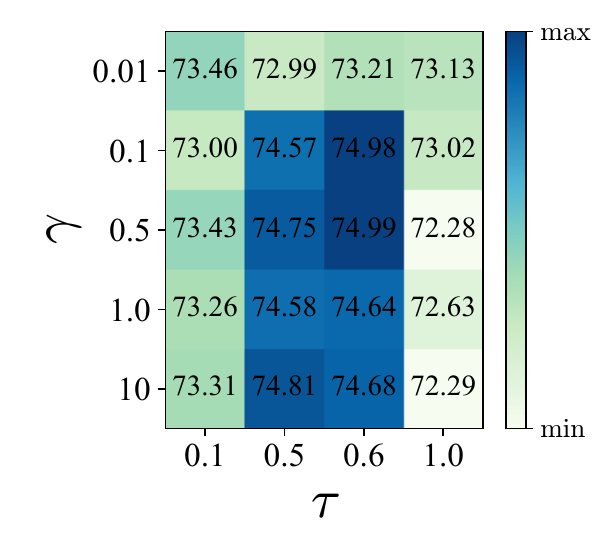}  
    \end{minipage} \label{figure_mcmix_gamma_tau_avg}   
    }\hspace{-12.1mm}              
    \subfigure[Last  accuracy ($\%$).]{
    \begin{minipage}[t]{0.263\textwidth}    
    \centering  \includegraphics[width=1\columnwidth]{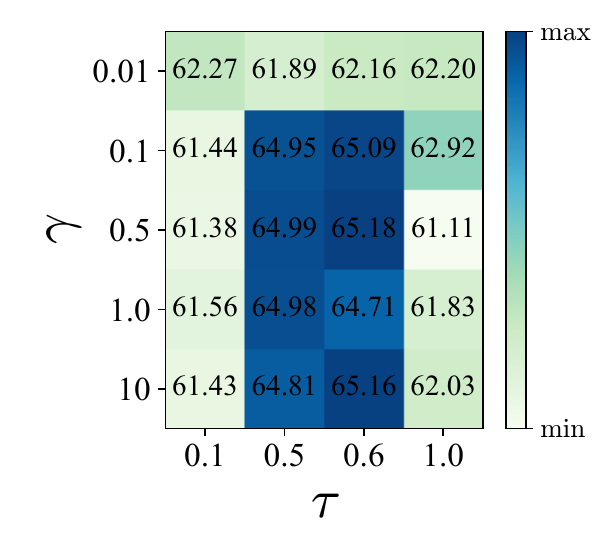} 
    \end{minipage}  
\label{figure_mcmix_gamma_tau_last}  
}\hspace{-11mm}
\end{center}    
\caption{Ablation comparison of hyper-parameters $\gamma$ and $\tau$ in MC-Mix on CIFAR-100 benchmark. }
\label{plot_mc_mix_gamma_tau}
\end{figure}

\subsection{Ablation Study}
\label{section_ablation_study}

In this subsection, we conduct an exhaustive ablation study to validate the effectiveness of our proposed method. We begin by presenting the main results and providing a component-wise analysis of the  ablation study. Then, we delve into a detailed analysis of each component.

\textbf{Main results of the ablation study.} Table~\ref{table_main_ablation_comparision} presents a comprehensive overview of the main results of our ablation study experiments. The CNN accuracy is directly determined by  utilizing the classifier head for prediction. In contrast, the nearest-mean-of-exemplars (NME) accuracy~\cite{DBLP:conf/cvpr/RebuffiKSL17} assigns class labels based on the most similar class feature prototypes. Intuitively, the CNN accuracy serves as an essential indicator for assessing the classifier's performance, while NME accuracy can offers valuable insights into the quality of the extracted feature representations.
Therefore, we systematically assess the individual impact of each component on both CNN accuracy and NME accuracy in the CIFAR-100 B0 10 steps and B50 5 steps benchmarks.

Taking into consideration the exclusive utilization of the dynamic architecture, as discussed in Section~\ref{section_dynamic_architecture}, within the baseline model, while excluding the incorporation of the MC-Mix strategy, vMF classifier, and matching loss, the baseline on CIFAR-100 B0 and B50 benchmarks achieve  average accuracies of $71.67 \%$ and $68.52 \%$, respectively. These results validate the considerable potential of the dynamic architecture approach in comparison to previous methods employing fixed-model size knowledge distillation techniques. Then, by introducing MC-Mix to the baseline model, the CNN accuracy is significantly improved by $+3.19 \%$, indicating its efficacy in boosting model performance, especially on the CIFAR-100 B0 10 steps benchmark.  Nevertheless, the observed improvements in NME accuracy is not significant enough, implying that the learning of feature representations may still be incomplete, emphasizing the necessity for further investigation and refinement of the feature learning for better performance.   Next, following the substitution of the original base classifier with the vMF classifier, notable performance gains are observed, particularly in terms of NME accuracy. Specifically, on the B0 10 steps benchmark, the NME accuracy increases from $72.24 \%$ to $73.69 \%$, and on the B50 5 steps benchmark, the NME accuracy improves from $68.69 \%$ to $69.50 \%$. These observations provide empirical validation of the substantial performance improvements achieved through the utilization of the vMF classifier and corroborate our analysis of using vMF classifier for dynamic feature learning in section~\ref{section_learning_dynamic_features_vmf}. Finally, we employ the proposed dynamic feature matching loss to synergistically align the vMF classifier and the dynamic feature representations. As for the B0 10 steps, we achieve an average accuracy of about $74.92 \%$ and $74.50 \%$ for both CNN and NME predictors, respectively. While the CNN accuracy remains relatively stable on B0 setting, it is noteworthy that NME exhibits a gradual improvement in accuracy over time, ultimately reaching parity with CNN accuracy, and we empirically use MC-Mix as model on ImageNet-1000 B0 10 steps setting in this case directly. Under the setting of B50 5 steps, we finally improve the CNN accuracy from $67.79 \%$ to $70.90 \%$ ($+3.21 \%$) compared to the baseline result, which validates the effectiveness of our method. In the remainder of this subsection, we proceed with a detailed analysis of the three components.    

\begin{table}[t]  
\caption{Ablation comparison of base classifier and vMF classifier on CIFAR-100 benchmark.} 
\label{table_base_classifier_vmf_classifier}
\begin{center}
\begin{tabular}{c|c|cc} 
\toprule
\multicolumn{2}{c|}{Benchmark} & \multicolumn{2}{c}{CIFAR-100 B50 5 steps}  \\ 
\midrule 
\multicolumn{2}{c|}{Classifier} & $\,\,\,\,$ CNN $\,\,\,$ & $\,\,\,\,$ NME  $\,\,\,$  \\ 
\midrule  
\multirow{2}{*}{{  $\textit{w}/\textit{o}$   MC-mix}} & $\,\,\,\,\,$ base $\,\,\,\,\,$ & $67.79$ & $68.52$ \\   
 & vMF  & $63.01$ & $67.20$  \\ 
\midrule  
\multirow{2}{*} {$\textit{w}/$ MC-mix} & base   & $67.86$ & $68.69$ \\
 & vMF   & $69.19$ & $\mathbf{69.50}$ \\   
\bottomrule  
\end{tabular} 
\end{center}
\end{table}

\begin{table}[t] 
\caption{Ablation comparison of mix-based data augmentation methods on CIFAR-100 benchmark.}  
\label{table_mix_based_data_aug}  
\begin{center}
\begin{tabular}{c|c|ccc}  
\toprule
 \multicolumn{2}{c|}{Benchmark} & \multicolumn{3}{c}{CIFAR-100 B0 - 10} \\ 
\midrule  
Methods & $\lambda(e)$  & \,\,\,\,\, Avg \,\,\,\,\, & Last & $\Delta$ \\   
\midrule 
- & - & $71.82$ & $60.11$ & - \\ 
Mixup & S & $69.20$ & $57.42$ & $-2.26/$$-2.69$  \\    
Cutmix & S & $72.41$ & $61.58$ & $+0.59/$$+1.47$ \\    
Cutmix ($w$)  & S & $72.86$ & $62.03$ & $+1.04/$$+1.92$ \\       
MC-Mix  & NS & $\mathbf{74.73}$ & $\mathbf{64.85}$ &   $\mathbf{+2.91/}$$\mathbf{+4.74}$ \\ 
\bottomrule 
\end{tabular} 
\end{center}
\end{table}


\textbf{Ablation study on MC-Mix (Hyper-parameters $\gamma$ and $\tau$).} 
Regarding the MC-Mix method, in addition to assessing its effectiveness as demonstrated in Table~\ref{table_main_ablation_comparision}, it also involves two hyper-parameters, namely $\gamma$ and $\tau$, which are related to non-stationary sigmoid function and require further analysis and sensitive study. As depicted in Fig~\ref{figure_sigmoid_functions_gamma_tau}, the parameter $\gamma$ regulates the steepness of function variation, where a larger $\gamma$ value leads to more pronounced and rapid changes, while a smaller $\gamma$ value results in smoother and gradual transitions. Conversely, $\tau$, ranging from $0$ to $1$, controls the positioning of non-stationary state changes within the sequence. To evaluate the hyper-parameters impact, we conduct a systematic investigation by exploring a range of values for $\gamma$ ($0.01, 0.1, 0.5, 1.0, 10$) and $\tau$ ($0.1, 0.5, 0.6, 1.0$) on CIFAR-100 B0 10 steps benchmark as shown in Fig~\ref{plot_mc_mix_gamma_tau}.

Firstly, when $\tau$ deviates from the central position and assumes values of $0.1$ and $1.0$, the model's performance experiences certain limitations. Although a smoother $\gamma$ (such as $0.01$) can partially mitigate these limitations with the last accuracy increasing from $61.43 \%$ to $62.27 \%$, the model still falls short of attaining the optimal results even with such refinements. Furthermore, we find that the hyper-parameter combination of $\gamma=0.5$ and $\tau=0.6$ consistently yields the best performance with average accuracy $74.99 \%$ and last accuracy $65.18 \%$. Notably, the favorable and similar performance tends to cluster around this particular parameter configuration. The robustness and insensitivity in hyper-parameter selection highlight the practicality of the proposed MC-Mix, which suggests that it can be seamlessly plug-and-play integrated into existing CIL methods as a data augmentation strategy. In this paper, we therefore set the parameter values of $\gamma$ and $\tau$ to $0.5$ and $0.6$, respectively.  

\begin{figure}[t]
\begin{center}
\includegraphics[width=0.28\textwidth]{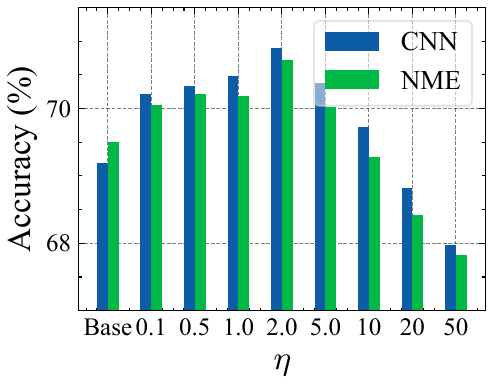} 
\end{center}  
\caption{Ablation comparison of hyper-parameter $\eta_{\mathrm{Ma}}$ in dynamic matching loss on CIFAR-100 benchmark.} 
\label{fig_plot_matching_loss_eta} 
\end{figure}   

\textbf{Ablation study on MC-Mix (Other mix-based methods).} On the other hand, as MC-Mix serves as a variant of mix-based data augmentation methods applied in CIL, it is essential to compare it with other mix-based approaches to validate its performance. Specifically, we include Mixup~\cite{DBLP:conf/iclr/ZhangCDL18}, CutMix~\cite{DBLP:conf/iccv/YunHCOYC19}, and CutMix with class weight information (referred to as Cutmix ($w$)) as the comparative methods for a  comprehensive analysis. As shown in Table~\ref{table_mix_based_data_aug}, we compared the performance of these three strategies, along with the baseline, against our proposed MC-Mix method on the CIFAR-100 B0 10 steps benchmark.

It is worth noting that the adoption of the Mixup results in a decline in model performance, whereas the utilization of CutMix exhibits a slight improvement. This observation reveals the significance of employing the CutMix regularization approach in preserving local image information with dynamic architecture. In line with this, our proposed MC-Mix method also builds upon the principles of CutMix. Moreover, the inclusion of normalized inverse class frequency weight information in CutMix mitigates the model bias caused by the imbalance between previous memory samples and new task samples, resulted in improved model performance with an average accuracy of $72.86\%$ and a last accuracy of $62.03\%$. However, the global stationary sequence of $\lambda(e)$ during training process still imposes limitations on the model's performance. The experimental results in  Table~\ref{table_mix_based_data_aug} illustrate that our proposed MC-Mix exhibits the best performance among the compared strategies, with an average accuracy of $74.73 \%$ and a last accuracy of $64.85 \%$. The average and last accuracies improved by $+2.91 \%$ and $+4.74 \%$ respectively compared to the baseline, which validates its effectiveness. Among the first three mix-based methods, the mean functions $\lambda(e)$ are assumed to be stationary sequences (S) by default, and CutMix ($w$) incorporates class weight information as an additional auxiliary component. However, MC-Mix takes a step further by introducing a non-stationary sequence (NS) with sigmoid functions and leveraging class weight information, leading to the ultimate and optimal improvement in performance.

\begin{figure}[t]
\begin{center}
    \subfigure[Baseline (example 1). ]{
    \begin{minipage}[t]{0.24\textwidth}   
    \centering     \includegraphics[width=1\textwidth] {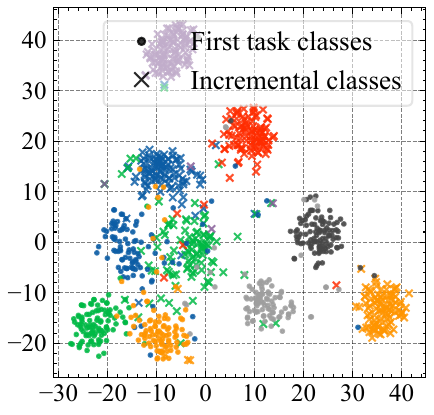}  
    \end{minipage} \label{figure_plots_tsne_der0}    
    }\hspace{-3.1mm}              
    \subfigure[Ours (example 1).]{
    \begin{minipage}[t]{0.24\textwidth}      
    \centering  \includegraphics[width=1\columnwidth]{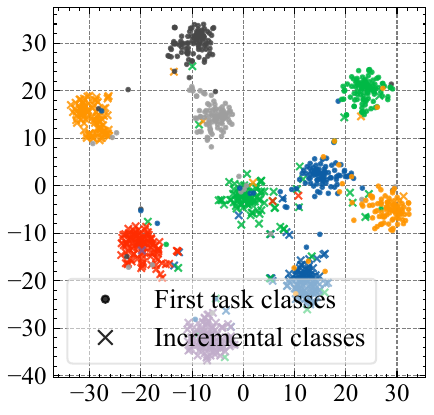} 
    \end{minipage}  \label{figure_plots_tsne_ours0}
    }\hspace{-3.1mm}
    \subfigure[Baseline (example 2). ]{
    \begin{minipage}[t]{0.24\textwidth}   
    \centering     \includegraphics[width=1\textwidth] {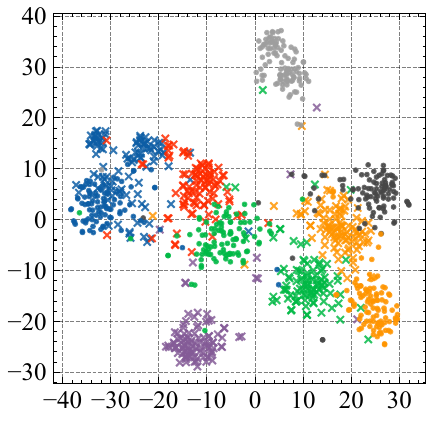}   
    \end{minipage} \label{figure_plots_tsne_der3}   
    }\hspace{-3.1mm}              
    \subfigure[Ours (example 2).]{
    \begin{minipage}[t]{0.24\textwidth}      
    \centering  \includegraphics[width=1\columnwidth]{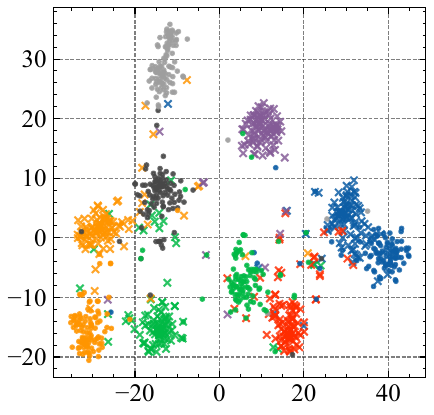} 
    \end{minipage}  \label{figure_plots_tsne_ours3}  
    }\hspace{-3.1mm}
\end{center}    
\caption{The two t-SNE feature visualization examples on CIFAR-100 benchmark. }
\label{plot_tsne_der_and_ours} 
\end{figure}

\textbf{Ablation study on vMF classifier.} We conducted a comparative analysis of the performance of the vMF classifier. As shown in Table~\ref{table_base_classifier_vmf_classifier}, we included an additional comparison of the performance using only the vMF classifier without MC-Mix on CIFAR-100 B50 5 steps benchmark. As described in section~\ref{section_learning_dynamic_features_vmf}, employing the vMF classifier enables better learning of dynamic features. However, we observed a discernible reduction in performance when exclusively substituting the original base classifier with the vMF classifier, resulting in CNN accuracy of $63.01 \%$ and NME accuracy of $67.20 \%$. We speculate that this could be attributed to the unit hyperspherical mapping by vMF classifier. The estimate of the mean direction weight is affected by the difference in sample size between the old and new samples. From another perspective, the NME accuracy of $63.01 \%$ improves over $3$ percentage points compared to CNN accuracy. This observation exemplifies the outstanding capability of the vMF classifier in effectively optimizing dynamic features, despite being impacted by classifier weight defects, which also lends support to our analysis framework and strengthens its credibility.

On the contrary, the utilization of the MC-Mix technique unleashes the latent potential of the vMF classifier method, leading to a notable improvement from the base classifier's performance of $67.86 \%$ and $68.69 \%$ to $69.19 \%$ and $69.50 \%$, providing empirical evidence supporting the validity of the vMF classifier. Particularly, the integration of the vMF classifier with MC-Mix yielded a significant improvement in CNN accuracy, increasing from $63.01 \%$ to $69.19 \%$ ($+6.18 \%$), which also demonstrates the success of the proposed MC-Mix approach.

\textbf{Ablation study on dynamic matching loss.} The dynamic matching loss also involves a hyper-parameter $\eta_{\mathrm{Ma}}$ to control the magnitude of the loss. Therefore, we further conduct analysis and sensitive study on CVIFAR-100 B50 5 steps with different $\eta_{\mathrm{Ma}}$ ($0.1, 0.5, 1.0, 2.0, 5.0, 10, 20, 50$), as shown in Fig~\ref{fig_plot_matching_loss_eta}.

Overall, different from the base model, when the matching loss is employed, it is clear that CNN outperforms NME in accuracy, regardless of different $\eta_{\mathrm{Ma}}$ magnitude. The collaborative alignment between the dynamic feature representations and the classifier leads to a subsequent unleashing of CNN accuracy thanks to the excellent features obtained by the vMF classifier. By setting $\eta_{\mathrm{Ma}}$ only to $0.1$, we can already observe performance improvement using a lightweight matching loss. When $\eta_{\mathrm{Ma}}$ is set too large, such as $20$ or $50$, there is a noticeable decline in performance compared to the base model. The excessively large matching loss overshadowed the utility of the cross-entropy loss for classification. Ultimately, the best results is obtained when $\eta_{\mathrm{Ma}}$ is set to $2.0$, yielding an accuracy of about $70.90 \%$, and we therefore set $\eta_{\mathrm{Ma}}$ to $2.0$ in this paper.

\begin{table}[!t]
\caption{Performance comparison using Plug and Play Memory-Centric Mix on the CIFAR-100 benchmarks.}
\label{table_plug_and_play_icarl_and_wa}
\begin{center}
\begin{tabular}{l|ccc} 
\toprule
Dataset        & \multicolumn{3}{c}{CIFAR-100}           \\ 
\midrule
Methods & B0 5 steps & B0 10 steps & B0 20 steps  \\ 
\midrule
iCaRL~\cite{DBLP:conf/cvpr/RebuffiKSL17} & $67.52_{\pm 1.45}$ & $64.06_{\pm 1.41}$ & $62.30_{\pm 1.48}$  \\  
iCaRL~\cite{DBLP:conf/cvpr/RebuffiKSL17} $\textit{w}/$ MC-mix & $69.78_{\pm 1.14}$ & $67.07_{\pm 1.38}$ & $64.92_{\pm 1.04}$  \\   
\midrule 
WA~\cite{DBLP:conf/cvpr/ZhaoXGZX20} & $69.15_{\pm 1.00}$ & $67.21_{\pm 1.34}$ & $65.31_{\pm 1.56}$  \\
WA~\cite{DBLP:conf/cvpr/ZhaoXGZX20} $\textit{w}/$ MC-mix  & $70.09_{\pm 0.94}$ &  $68.27_{\pm 1.06}$ & $65.38_{\pm 1.60}$  \\
\bottomrule
\end{tabular}
\end{center} 
\end{table}

\begin{figure}[t]
\begin{center}
    \subfigure[iCaRL. ]{
    \begin{minipage}[t]{0.24\textwidth}   
    \centering     \includegraphics[width=1\textwidth] {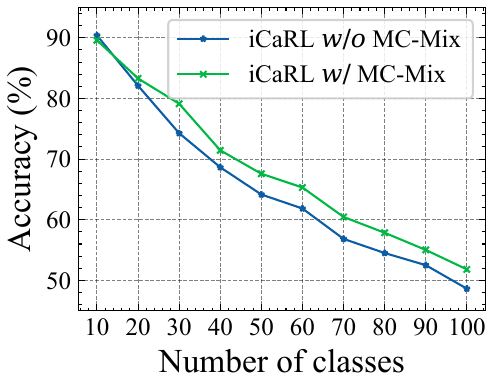}  
    \end{minipage} \label{figure_plot_plug_and_play_icarl}   
    }\hspace{-3.1mm}              
    \subfigure[WA.]{
    \begin{minipage}[t]{0.24\textwidth}     
    \centering  \includegraphics[width=1\columnwidth]{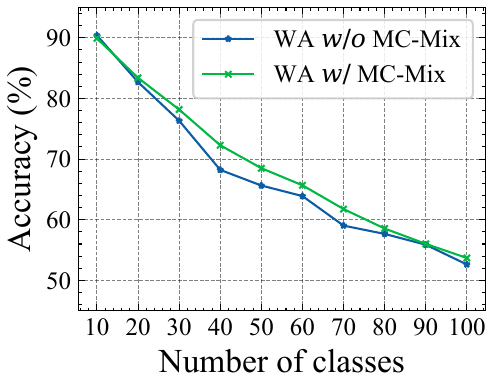} 
    \end{minipage}  
\label{figure_plot_plug_and_play_wa}  
}\hspace{-11mm}
\end{center}    
\caption{Performance comparison for each step using Plug and Play Memory-Centric Mix on the CIFAR-100 benchmark. }
\label{plot_plug_and_play_icarl_and_wa}
\end{figure}

\begin{figure*}[t]
\centering    
\subfigure[Linear function ($w_y=0.5$).]{
    \begin{minipage}[t]{0.246\textwidth}
    \centering     \includegraphics[width=1\textwidth]{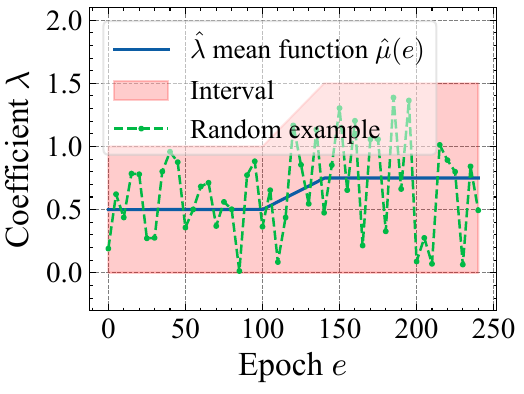} 
    \end{minipage} 
\label{figure_plot_ns_linear_curve_old} 
}\hspace{-3mm}  
\subfigure[Linear function ($w_y=-0.5$).]{
    \begin{minipage}[t]{0.246\textwidth}
    \centering
    \includegraphics[width=1\columnwidth]{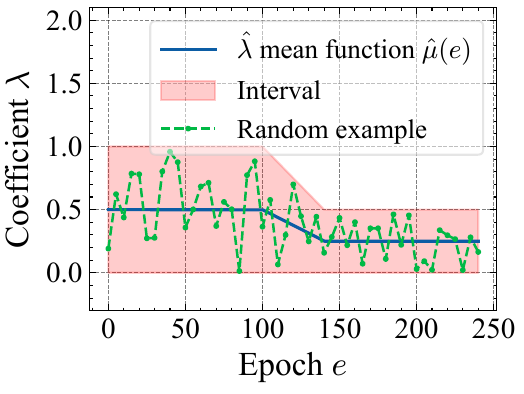} 
    \end{minipage}  
\label{figure_plot_ns_linear_curve_new}   
}\hspace{-3mm}    
\subfigure[Step function ($w_y=0.5$).]{
    \begin{minipage}[t]{0.246\textwidth}
    \centering 
    \includegraphics[width=1\columnwidth]{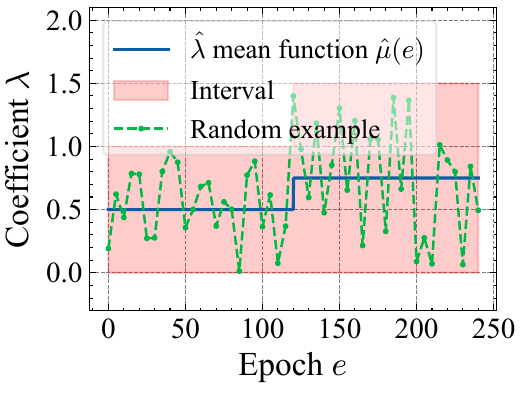} 
    \end{minipage}
\label{figure_plot_ns_step_curve_old}
}\hspace{-3mm} 
\subfigure[Step function ($w_y=-0.5$).]{
    \begin{minipage}[t]{0.246\textwidth}  
    \centering \includegraphics[width=1\columnwidth]{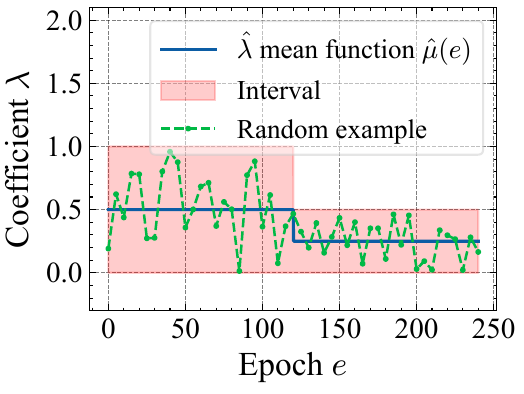} 
    \end{minipage}
\label{figure_plot_ns_step_curve_new}   
}  
\caption{Other Non-stationary Sequence Functions in MC-Mix.   }
\label{plot_other_non_stationary_function_mean_function}
\end{figure*}   

\subsection{Further Exploration}
\label{subsection_further_exploration}

In this section, we conduct an in-depth investigation of our methodology, including analysis of feature t-SNE visualization~\cite{van2008visualizing}, seamless integration of MC-Mix with other CIL techniques, and the exploration of other non-stationary functions.

\textbf{t-SNE Feature  Visualization}. We employ t-SNE~\cite{van2008visualizing} to visually analyze the dynamic feature representations of the CIFAR-100 test dataset. Fig.~\ref{plot_tsne_der_and_ours} present two comparative results of the baseline and our proposed model on the CIFAR-100 B50 5 steps evaluation setting. Specifically, distinct marker symbols and colors are utilized to symbolize different class samples. For better  visualization, we select five classes from first task and mark them with circles "$\circ$", and in each of the following five incremental tasks, select one class and mark with "$\times$". Therefore, we  ultimately visualize dynamic features samples from differently selected 10 classes with two examples.

For example 1 shown in Fig.~\ref{figure_plots_tsne_der0} and Fig.~\ref{figure_plots_tsne_ours0}, the five categories located in the lower left corner of the baseline results, namely blue $\circ$, blue $\times$, green $\circ$, green $\times$, and orange $\circ$, exhibit a tendency towards dense clustering. On the contrary, our method shows a tendency to increase the inter-class distance, especially for green $\circ$ and orange $\circ$. As for example 2 shown in Fig.~\ref{figure_plots_tsne_der3} and Fig.~\ref{figure_plots_tsne_ours3}, it is worth noting that the the class of blue $\times$  is split into two parts in baseline. However, as for ours results, although there is a slight overlap between the blue $\circ$ and $\times$ in Fig.~\ref{figure_plots_tsne_ours3}, it is evident that the blue  $\times$ instances belonging to the same class exhibit noticeable aggregation. These observed results validate that our model learn better dynamic features, confirming its ability to resist catastrophic forgetting in class-incremental learning. Moreover, these visualization results  corroborate the improvement in NME accuracy observed in ablation study and are also consistent with our vMF classifier analysis, further validating the superiority of our method.

\begin{figure}[t]
\begin{center}
    \subfigure[Average accuracy ($\%$). ]{
    \begin{minipage}[t]{0.239\textwidth}     
    \centering     \includegraphics[width=1\textwidth] {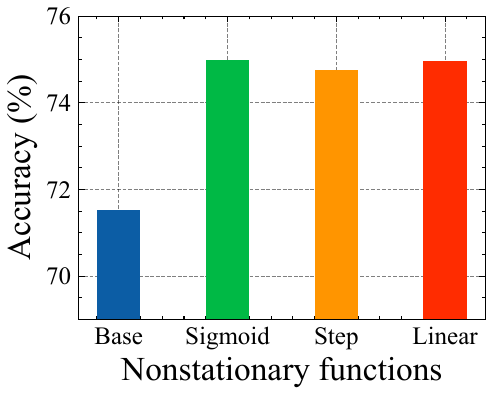}  
    \end{minipage} \label{plot_mc_mix_other_nonstationary_functions_avg}   
    }\hspace{-3mm}                
    \subfigure[Last  accuracy ($\%$).]{
    \begin{minipage}[t]{0.239\textwidth}        
    \centering  \includegraphics[width=1\columnwidth]{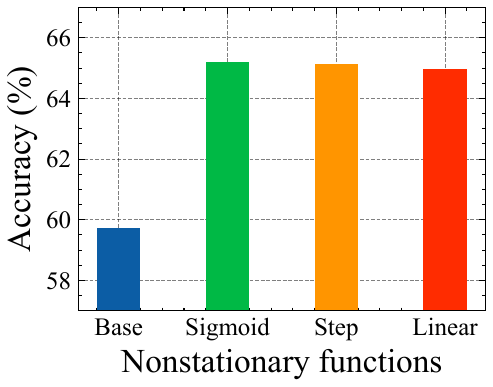} 
    \end{minipage}  
\label{plot_mc_mix_other_nonstationary_functions_last}  
}\hspace{-11mm}
\end{center}    
\caption{Ablation comparison of different non-stationary functions in MC-Mix on CIFAR-100 benchmark. } 
\label{plot_mc_mix_other_nonstationary_functions}
\end{figure}

\textbf{Plug-and-play MC-Mix.} As stated in the related work in section~\ref{section_related_work_da_in_cil}, the data augmentation strategies intertwine with the techniques of CIL. Hence, our proposed MC-Mix method offers a seamless and plug-and-play integration with existing CIL techniques, allowing for its immediate utilization. As shown in Table~\ref{table_plug_and_play_icarl_and_wa}, we consider two classic CIL methods, namely iCaRL~\cite{DBLP:conf/cvpr/RebuffiKSL17} and WA~\cite{DBLP:conf/cvpr/ZhaoXGZX20}, and applied MC-Mix to them in order to further improve the performance. The benchmark comparisons are performed on the CIFAR-100 dataset B0 5 steps, 10 steps, and 20 steps. We also run experiments using three different class orders and report their mean and standard deviation.

Overall, the experimental results in Table~\ref{table_plug_and_play_icarl_and_wa} indicate that the average accuracy achieved using MC-Mix consistently surpasses the accuracy obtained without MC-Mix. Particularly, for iCaRL method, we can clearly observe a significant improvement from $64.06 \%$ to $67.07 \%$ ($+3.01 \%$) in the average accuracy on B0 10 steps benchmark. On the other hand, the average accuracy of WA exhibits a modest increase from $67.21 \%$ to $68.27 \%$ ($+ 1.06 \%$), as compared to the improvement observed in iCaRL. We speculate that this difference can be attributed to the fact that WA mitigates the issue of sample imbalance between new and old tasks by aligning classifier weights, which is partially coupled with the utility of our MC-Mix methodology. Nevertheless, we indeed observe a slight boost, providing empirical evidence for the effectiveness of MC-Mix. Fig.~\ref{plot_plug_and_play_icarl_and_wa} presents the performance comparison at each step using the MC-Mix on the CIFAR-100 benchmark, we can see a consistent improvement in accuracy for almost every task, further proving the success of the MC-Mix.

\textbf{Other Non-stationary Functions.} One of the key components of our proposed MC-Mix is the selection of non-stationary functions as shown in Fig.\ref{plot_sigmoid_function_mean_function}. The insight behind our choice of the sigmoid function is rooted in its common usage as a non-linear activation function. In this subsection, we further explored other non-stationary functions, including the unit step function and linear function. It is worth noting that the tanh function can be directly considered as a variant of the sigmoid function. Similar to Fig.~\ref{figure_sigmoid_function_old_weight} and Fig.~\ref{figure_sigmoid_function_new_weight}, we also provide the plots of mean functions $\hat{\mu}(e)$ corresponding to the linear function and the unit step function as shown in Fig.~\ref{plot_other_non_stationary_function_mean_function}. It is clear that the class weight information is 
introduced to dynamically adjust the distribution range of the interpolation coefficients $\lambda$. For quantitative analysis, we conducted comparative evaluation between the baseline model and other non-stationary functions on CIFAR-100 B0 10 steps benchmark as shown in Fig.~\ref{plot_mc_mix_other_nonstationary_functions}.

The experimental results demonstrate a significant improvement in both average accuracy and last accuracy of the baseline model when non-stationary functions are used. It is noteworthy to observe that the utilization of different functions did not exhibit significant difference, thereby attesting to the robustness of MC-mix. Furthermore, we conjecture that epoch as a natural number results in discrete sampling of different non-stationary functions that do not vary much during model training.  Nevertheless, it is imperative to emphasize that empirical evidence strongly supports the indispensability of incorporating non-stationary mean functions for performance improvement as demonstrated in Table~\ref{table_mix_based_data_aug}.

\section{Conclusion}

In this paper, we improve the dynamic architecture based CIL model and propose our Dynamic Feature Learning and Matching (DFLM) model from three perspectives: data augmentation, feature representation, and classifier. Memory-Centric Mix strategy is proposed that dynamically attends to memory replay samples by incorporating non-stationary class weights information. Then, vMF classifier is introduced to effectively model and learn the dynamic features. The matching loss finally aligns the classifier weights with dynamic features by minimizing distribution distance. Extensive experiments show that our model obtains superior performance on CIL benchmarks.


%

\ifCLASSOPTIONcaptionsoff
  \newpage
\fi



\bibliographystyle{IEEEtran}
\bibliography{references}
%

%

\end{document}